# Sketch2Sim: an end-to-end modular framework for hand-drawn circuit analysis

Mohammad Kawsar[1], Al Amin Islam Ridoy[1], Md. Rakib Hossain[1], Farzana Akter[1], AKM Bahalul Haque[2], Rejwan Bin Sulaiman[3] and Mohammed Amin Almaiah[4]

[1] Department of Engineering & Technology, Rajshahi University of Engineering & Technology, Rajshahi, Bangladesh
[2] Department of Information Technology, Åbo Akademi University, Turku, Finland
[3] Computer Science and Technology, University of Northumbria at Newcastle, London, United Kingdom
[4] The King Abdullah II School of Information Technology, The University of Jordan, Amman, Jordan

## ABSTRACT

**Context:** The digitization of handwritten circuits plays a crucial role in the education and engineering domains. They can help analyze circuits and provide feedback that enables students, researchers, educators, and other stakeholders. Additionally, due to the spread of digital platforms in education, handwritten circuit digitization can help students and educators in various developing countries who lack access to fully digitized content. However, this topic has remained underexplored due to data scarcity, lack of datasets, and the complexity of processing.

**Objective:** This study proposes a sketch-to-simulation system to solve handwritten circuit diagrams. The system converts raw images into a Simulation Program with Integrated Circuit Emphasis (SPICE)-compatible netlist and automatically calculates the nodal voltages and currents of the circuit. Another objective is to propose a holistic framework that benefits students and educators by enhancing the teaching and learning experience of electric circuits and expediting engineering design reviews.

**Methods:** This article presents a multistage pipeline that converts raw photographs of handwritten electric circuits into executable SPICE-like netlists, providing a solution. A dataset of 1,000 hand-drawn Resistor-Inductor-Capacitor (RLC) circuits was curated and annotated for components, text regions, and character recognition. The proposed system decomposes a circuit diagram into three stages: (i) dual You Only Look Once (YOLO)-based detectors for components and text regions, (ii) a Convolutional Recurrent Neural Network (CRNN) with Connectionist Temporal Classification (CTC) and Optical Character Recognition (OCR)-based domain-aware symbol value extraction, and (iii) image processing with thresholding, skeletonization, and Density-Based Spatial Clustering of Applications with Noise (DBSCAN) to automatically identify nodes and map them with corresponding components.

**Results:** The proposed system produces the netlist and solves the circuit. The qualitative and quantitative ablation studies on modular stages, as well as end-to-end tests, demonstrate robust performance across various handwriting styles, lighting conditions, and clutter.

**Conclusion:** This study serves as an interactive learning aid and practical tool for professionals, providing a comprehensive user interface (UI), enabling rapid digitization of circuit sketches for study, design, prototyping, or troubleshooting.

Submitted 24 December 2025
Accepted 19 August 2026
Published 13 October 2026

Corresponding author
Rejwan Bin Sulaiman,
rejwan.sulaiman@northumbria.ac.uk

## INTRODUCTION

Hand-drawn circuit diagrams are the go-to medium for brainstorming ideas, teaching basics, or rapid design reviews. They play a critical role by showing how electric components are connected in real-life applications (*Yang et al., 2024*), from new buildings and factories to power plants that light up entire cities (*Xie et al., 2022*; *Theisen et al., 2023*). Yet, manual processing or tuning a sketch on article into simulation-ready format demands tedious redrawing in Computer-Aided Design (CAD)/Electronic Design Automation (EDA) tools, by hand re-entry of values and wiring. Such steps are exacting, slow, and error-prone (*Ekeke, Moreno-García & Elyan, 2025*; *Wang et al., 2025*; *Daele et al., 2021*). This, in turn, drives the demand for computer-vision-based frameworks that can automatically process circuit diagrams (*Jamieson, Moreno-García & Elyan, 2024*; *Moreno-García, Elyan & Jayne, 2019*). Bridging the gap requires the development of robust pipelines that can yield a simulation-ready netlist for engines such as Simulation Program with Integrated Circuit Emphasis (SPICE) or that can produce a circuit solution (*Peker et al., 2026*).

The concept of automated circuit simulation is challenging for three reasons: component detection, text recognition, and topology extraction (*Hu, Zhan & Tong, 2024*). The components inside circuit diagrams are drawn in diverse styles; they are densely packed, vary in size, angles, and line widths. Moreover, near similar structure type components (*e.g.*, resistors and inductors) may appear visually the same due to hand-drawing, which makes them difficult to identify. Secondly, extracting text from circuit diagrams is also challenging due to mixed digits, letters, domain symbols, and often due to overlapping. On top of that, extracted values must be parsed numerically and normalized while preserving units. Finally, the wires of hand-written diagrams are noisy-curvy lines, over-drawn corners, and small gaps make it even harder to detect and assign nodes to the corresponding component. Prior systems tend to focus only on one of these aspects (*e.g.*, symbol detection or text recognition), but rarely deliver a fully automated end-to-end system for electric circuits without manual clean-up (*Peker et al., 2026*).

This article presents a sketch-to-simulation system designed to tackle the above challenges that automatically solves raw hand-written circuit diagrams. As there was no suitable dataset for analog Resistor-Inductor-Capacitor (RLC) networks, this article also proposed a dataset of 1,000 hand-written circuit diagrams drawn from different persons. To reduce the ambiguity in topology reconstruction, we primarily focus on hand-drawn 1st-order RLC circuits with independent sources. This design choice provides a controlled benchmark for validating the proposed end-to-end pipeline. However, the modular architecture of the framework allows extension to more complex circuit topologies in future work. Another three versions of the proposed dataset are created for component detection, text region detection, and optical character recognition (OCR) purposes.

Utilizing the curated dataset, the framework is created to automatically identify components, read component values, process network topology, generate simulation-ready netlist and finally performs nodal analysis to solve the circuit without human intervention *via* You Only Look Once-version 8 (YOLOv8) detectors, Convolutional Recurrent Neural Network (CRNN)-Connectionist Temporal Classification (CTC), and topology reconstruction *via* skeletonization and Density-Based Spatial Clustering of Applications with Noise (DBSCAN). Such a framework is beneficial both for educational and industrial settings-enhancing the teaching or learning experience of electric circuits and expediting engineering design reviews.

The contributions of the present work are concisely highlighted below:

1. A fully automated sketch-to-simulation pipeline that converts hand-drawn circuit diagrams into SPICE-compatible netlists and solves the circuit through nodal analysis.
2. A novel dataset of 1,000 handwritten circuit diagrams has been proposed, including annotations for component detection, text localization, and OCR of engineering values.
3. A hybrid framework combining deep learning and classical graph extraction, integrating YOLOv8 detectors, CRNN-CTC OCR, and topology reconstruction *via* skeletonization and DBSCAN.
4. Extensive experimental validations, ablation studies, and a deployable web interface enabling interactive circuit verification and simulation.

The latter part of the article is organized as follows: 'Background and Related Works' presents a review of related research on component detection, text extraction, and simulation steps involved in hand-written circuit automation. 'Methodology' provides a detailed methodology of the proposed framework. 'Result Analysis' discusses the results deduced from this research. After that, 'Limitations' describes some limitations of this proposed work, and 'Discussion' discusses the research and practical implications of this proposed framework. Finally, the overall article is summarized in 'Conclusion and Future Work' with potential developments for the future.

## BACKGROUND AND RELATED WORKS

Although hand-drawn circuit interpretation is an under-explored research area, some researchers have focused on individual sections of the targeted end-to-end task. The first step of automatic circuit interpretation is the detection of components. Early research on component detection relied upon pattern recognition and rule-based approaches. *Okazaki et al. (1988)* pioneered automating the reading of circuit diagrams by proposing symbol segmentation and identification through their closed-loop shapes. Later, *Feng, Viard-Gaudin & Sun (2009)* improved the component segmentation and recognition method by proposing a two-dimensional dynamic programming technique. However, the efficacy of earlier methods was limited by noise as well as complex component connectivity, and could not handle variations in symbol style.

Prior to the advancements of artificial intelligence, *Roy et al. (2020)* improved robustness to symbol style variation by extracting texture and shape-based features to classify hand-drawn circuit components with a classical machine learning pipeline. An extension of that work was introduced by *Dey et al. (2021)* where two staged CNN (*Lecun et al., 1998*) firstly clustered the similar looking component into one group, and then classified them into their true class. Although with the cost of speed, the two-staged framework yields better classification accuracy than single staged, there was a noticeable amount of error in clustering similar components. In this regard, object detection models such as Faster R-CNN (*Ren et al., 2015*; *Yang et al., 2024*) and You Only Look Once (YOLO) (*Redmon et al., 2016*) showed propitious performance (*Yang et al., 2023*).

Based on experimental comparison among YOLO, Faster R-CNN, and other deep learning methods on hand-written binary logic gate detection, *Amraee et al. (2022)* found that YOLO performed better than others. Upon detection, they successfully generated a binary function and truth table from logic circuit diagrams using a simple boundary tracking method. As the system is feasible for simple wiring, it suffers a lot on complex circuits. To analyze power converter circuits, *Bohara & Sarma Krishnamoorthy (2024)* also applied the YOLO model for hand-drawn component identification and utilized classical techniques for netlist creation. It achieved 95.17% accuracy on netlist generation when tested on external circuits.

Not only the component detection, but the corresponding value extraction of each component is equally significant. In early times, handwritten word recognition was addressed through the Hidden Markov Model (HMM) (*El-Yacoubi et al., 1999*). But hybrid models such as HMM + Neural Network (*Knerr & Augustin, 1998*) or HMM + Recurrent Neural Network (RNN) (*Senior & Robinson, 1995*) demonstrated better performance than HMM alone. These approaches were limited by low text discriminative power, and word error rates were higher. Similar to speech recognition, the development of an encoder-decoder-based framework demonstrated promising results in handwritten character recognition (*Bluche, 2016*; *Lee & Osindero, 2016*). Unlike prior methods that required explicit character or word-level segmentation (*Papavassiliou et al., 2010*), an encoder-decoder-based framework handles text holistically, without explicit segmentation (*Coquenet, Chatelain & Paquet, 2023*).

*Shi, Bai & Yao (2017)* proposed an attention-based encoder-decoder approach, where text inside the images is encoded as features and decoded as characters. The stacks of Convolutional Neural Network extracted the features, bidirectional long short-term memory (BiLSTM) modeled the sequence, and CTC (*Graves et al., 2006*) layer performed transcription to yield the final character sequence. Such two-meta algorithms were widely used in several approaches (*Li et al., 2019*; *Wang & Hu, 2017*), and our work's text recognition part is inspired by it. However, these literatures were targeted for English alphanumeric and suffers from or misjudges out-of-the-box circuit notation (*e.g.*, interpreting 10 Ω as 100).

Besides component detection and value extraction, recovering the connectivity graph from the circuit is often the most challenging aspect. *Rachala & Panicker (2022)* proposed a system where the Hough transform (*Mukhopadhyay & Chaudhuri, 2015*) was used to identify wire strokes, and junctions were inferred where multiple lines met. Such approaches may work well on straight connections but will struggle with hand-written curved or slightly jagged wires. Another study (*Lakshman Naika, Ramegowda & Soukar, 2019*) applied skeletonization combined with connectivity analysis to find line endpoints and intersections. Without noisy handling, such systems could produce spurious node candidates. After removing the component from the diagram, *Alhalabi et al. (2021)* treated wire blobs as a single node and assumed every component has two connection points. As a result, merging multiple junctions or branches will be combined as a single node, and endpoints or intersections will become indistinguishable. A graph neural network (GNN) can also be used to detect and extract complex network topologies (*Renton et al., 2021*).

Prior studies typically used a single detection model for symbols or texts, whereas the proposed system employed a dual detector- one specialized for component and the other for text region identification. This separation is aimed at improving precision on each task. Additionally, a detailed ablation study for model selection has been conducted, exploring the impact of hyperparameters, model size, and performance. Such systematic experiments are absent in earlier works, which most typically dwelt on reporting results for one chosen configuration. Moreover, the coupling of a dedicated text-region detector with the OCR model of this study is capable of decoding engineering symbols and interpreting unit suffixes (*e.g.*, 2.2 μF) correctly even in messy handwritten labels. This approach goes beyond the OCRs used in previous studies.

The application of thinning (*Lam, Lee & Suen, 1992*) algorithm to detect endpoints/intersections followed by DBSCAN clustering (*Kumar & Reddy, 2016*) to produce stable set of nodes by filtering noise made the proposed system capable of handling imperfect sketches, where previous studies would struggle (*Agrawal, Jagtap & Kantipudi, 2024*). This allows the proposed approach to extract the circuit graph correctly for the vast majority of test sketches, as described in 'Node Detection, Final Annotation, Netlist Generation, and Circuit Solution'. In contrast to prior studies that either just detect components (*Rica, Álvarez & Serratosa, 2021*; *Antonelli et al., 2022*; *Zhang et al., 2022*), extracts texts (*Chen et al., 2021*; *Kim et al., 2021*) or, in very few cases, produce labeled diagrams and a netlist (*Peker et al., 2026*; *Hu, Zhan & Tong, 2024*; *Hemker et al., 2024*), we propose a complete end-to-end system starting from uploading an image to the final nodal solution with a real-time web implementation. The web implementation enables humans to correct any errors that occur in any circuit interpretation step, which lowers the barriers of leveraging such technology.

## METHODOLOGY

This section presents the process used to develop the circuit simulation framework. From dataset creation to netlist generation and circuit solution, every step taken thus far is described below:

### Dataset

Real-world hand-written circuits include messy lines and component drawing, along with symbols or numeric values. Moreover, writing varies from person to person. Therefore, a dataset of natural diversity in handwriting styles, spacing, and symbol representation was necessary (*Ahmed et al., 2025*). Since there exists no suitable dataset for interpreting hand-written circuit diagrams and generating machine-readable netlists, a new dataset has been manually created from scratch in this study. Different electrical books have been followed to manually draw the circuit diagrams, and a total of 1,000 hand-written circuit diagrams drawn by different persons have been curated to build this dataset. The raw handwritten image dataset has been further processed. Each Red, Green, and Blue (RGB) image was converted to a gray-scale to remove color noise, *e.g*., pencil shades. Then the images were passed through a Gaussian along with non-local means filtering to reduce additional noise, but keeping the edges. After that, the images were cropped to eliminate the unnecessary borders, and the contrast of the circuit diagrams was enhanced. Some samples of the original dataset and their processed version, both are depicted in Fig. 1. Finally, the images were resized to 272 × 272 pixels. The task-ready processed image has undergone three stages to create the additional datasets, which are described in Fig. 2.

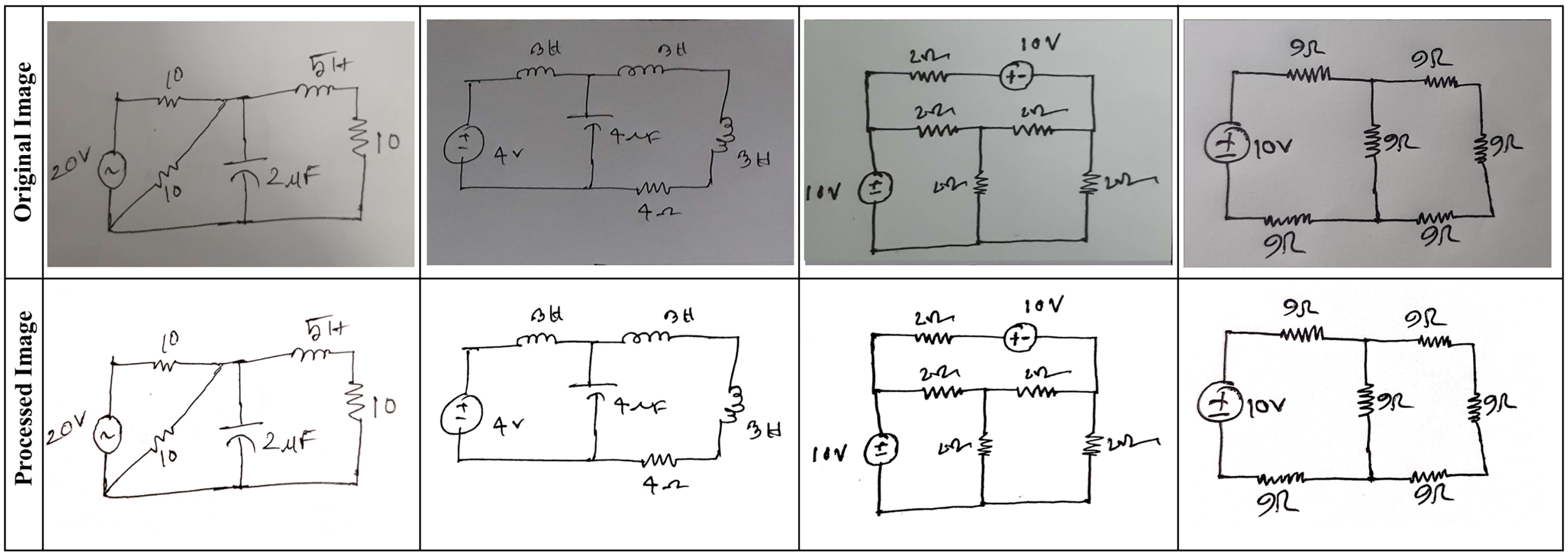

Figure 1 **Sample circuit images with annotations.** Full-size DOI: 10.7717/peerj-cs.4103/fig-1

### *Stage 1: component detection dataset*

The dataset consists of six types of components. As these are 1$^{st}$ order RLC circuits, the number of resistive elements is higher than the other components. 45.3% of the total components are resistors, whereas the other elements (inductor, capacitor, current source, AC, and DC voltage component) combinedly exhibits 54.7%. The exact class-wise component distribution of the dataset is presented in Fig. 3. The first task in the pipeline was to detect the fundamental component within a circuit. Therefore, each image of the dataset was annotated manually using the *LabelImg* tool (https://labelimg.org/), with bounding boxes drawn tightly around each component, as demonstrated in Fig. 4A. The annotations were saved in YOLOv8 format, specifying class IDs and normalized box coordinates. To maintain consistency, annotation guidelines were established and followed throughout the dataset labeling process. This included rules on handling overlapping symbols and minimum box margins.

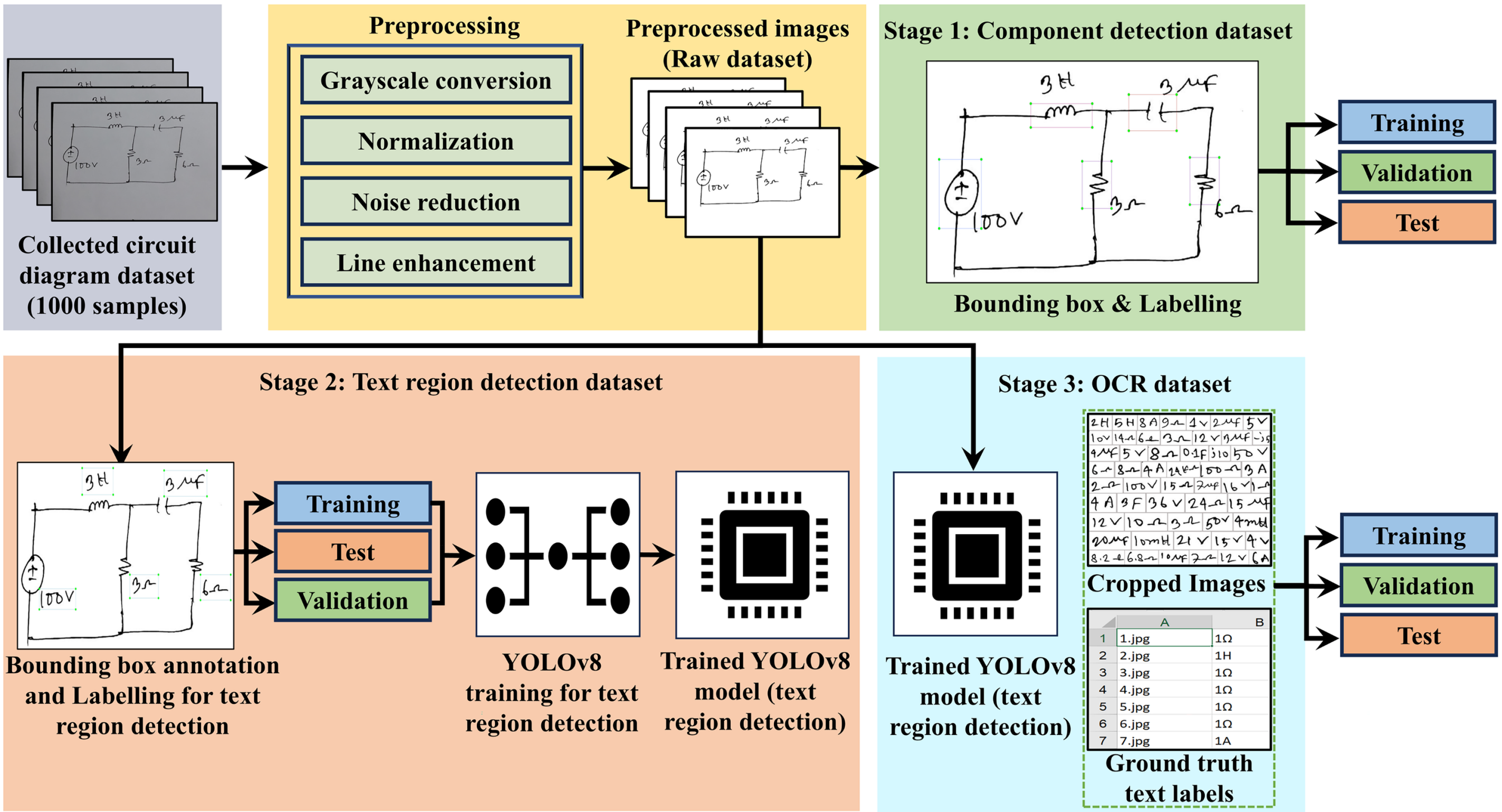

Figure 2 Dataset preparation workflow showing preprocessing, component annotation (six classes), text region annotation, and cropped OCR dataset creation. Full-size DOI: 10.7717/peerj-cs.4103/fig-2

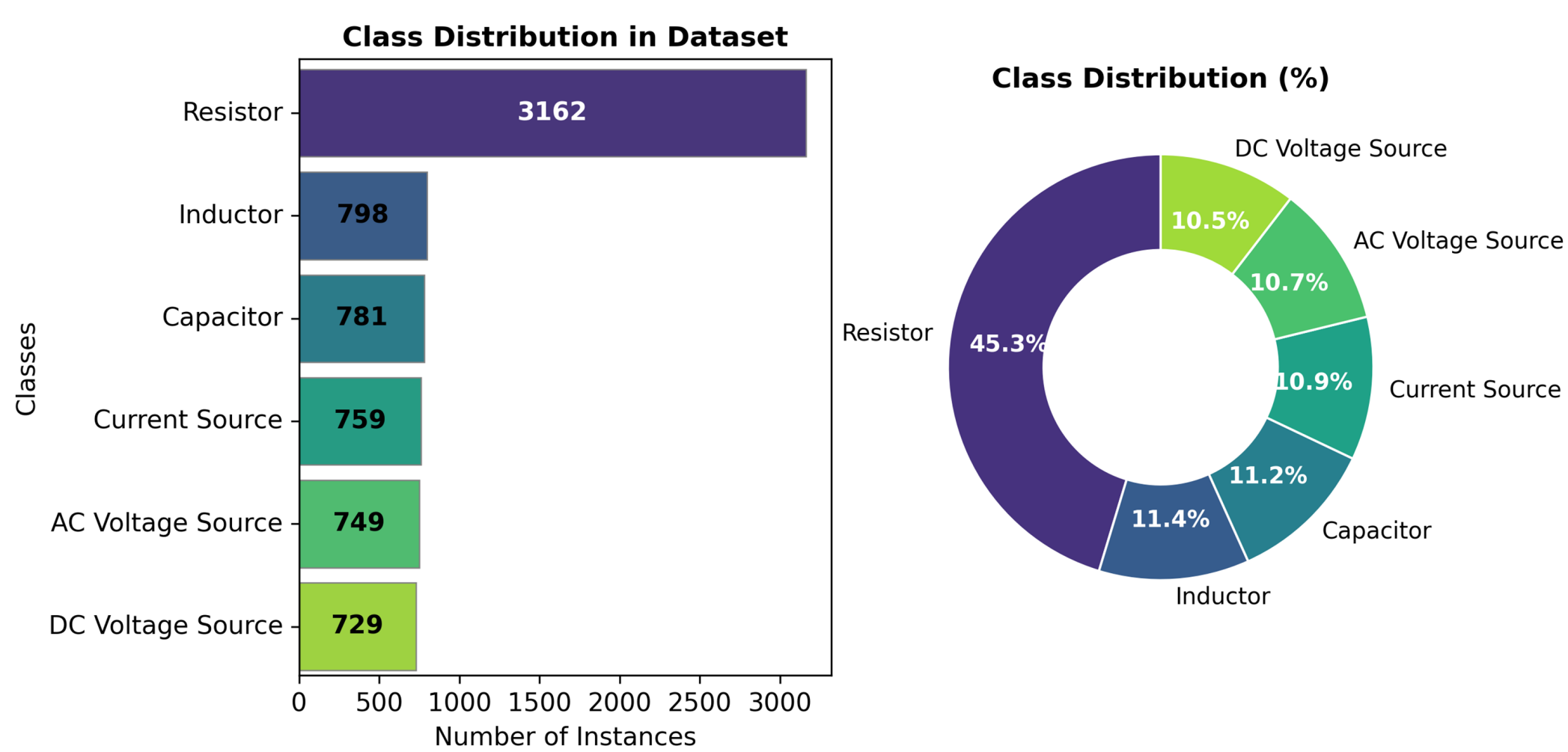

Figure 3 Class-wise distributions of the components of the entire hand-drawn electric circuit diagram dataset. Full-size DOI: 10.7717/peerj-cs.4103/fig-3

### *Stage 2: text region detection dataset*

In electric circuits, each component is accompanied by its unit value (*e.g.*, "10 Ω", "2 μF", "5 V"), which is basically a text. Hence, it is necessary to detect that text region. In addition to component detection, a second round of annotations was performed to localize these text regions. An example of text region annotation is depicted in Fig. 4B. The dataset contains 6,978 annotated component instances, of which approximately 3,000 contain handwritten value labels. Therefore, around 3,000 text regions were annotated for OCR training. After completing text region annotating, the Yolov8 model was trained on those images to make it specialized for identifying bounding boxes around any form of handwritten text, regardless of the content. This model has been used to crop the bounding box of the text region from the images, and these cropped images have been used to create the OCR dataset.

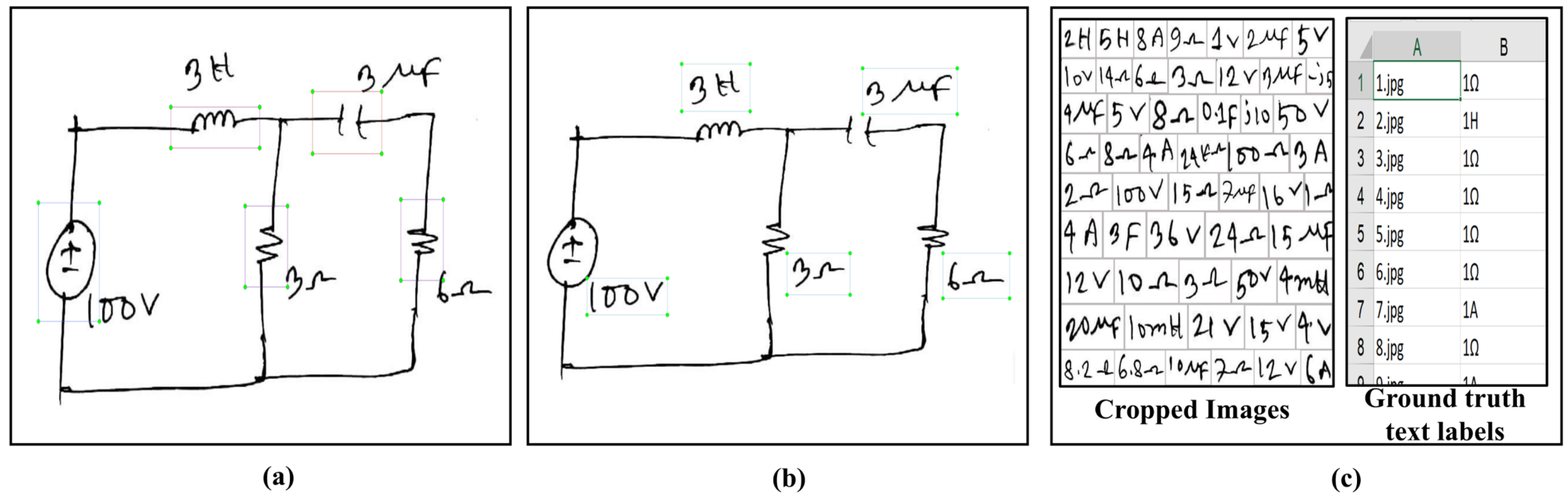

**Figure 4** **Preparation of (A) a component detection dataset with six types of component annotations. (B) A text region detection dataset containing only text annotations. (C) The final OCR dataset consisting of cropped text images and their corresponding image filenames with ground-truth text labels.** Full-size DOI: 10.7717/peerj-cs.4103/fig-4

**Table 1** **Summary of the datasets used in each stage.**

| Dataset type | Purpose | No. of images | Classes | Total component or instances | Split ratio (%) | | | Model used |
|---|---|---|---|---|---|---|---|---|
| | | | | | Train | Test | Validation | |
| Component detection | Detect six circuit elements | 1,000 | 6 | 6,978 | 70 | 20 | 10 | YOLOv8 |
| Text region detection | Localize handwritten values | 1,000 | 1 | 6,978 | 70 | 20 | 10 | YOLOv8 |
| OCR text dataset | Recognize values/units | 3,000 | 68-char | 3,000 | 70 | 20 | 10 | CRNN + CTC |

### *Stage 3: OCR dataset*

After cropping the text regions of the circuit diagrams, each cropped text image was saved and named in numeric order in JPG format. Additionally, each cropped text image's name and its corresponding text have been written in an Excel file, as shown in Fig. 4C. Image text of the OCR dataset includes a custom character set of 68 symbols: digits (0–9), uppercase and lowercase alphabets, and electrical symbols such as ohm (Ω), micro (μ), voltage (V), ampere (A), inductance (H), and capacitance (F). This extended character set was crucial for correctly interpreting engineering-specific units and values.

## Training process

Once the dataset is prepared, the next step is to train the deep learning models that form the foundation of the proposed system. As described in Table 1, 70% of the data of each dataset was used for training, 20% for testing, and 10% for validation purposes. For component and text region detection, two different YOLOv8 models were used, and for text extraction, a CRNN + CTC model was trained. While training the models, multiple YOLOv8 variants (nano, small, and medium) and different hyperparameter configurations were explored to choose the optimal model for each task. The extensive data-driven comparison is presented in 'Result Analysis' section. After fine-tuning with hyperparameters, the weights of the best models were saved for later use in the end-to-end pipeline. The overall training process is described in Fig. 5.

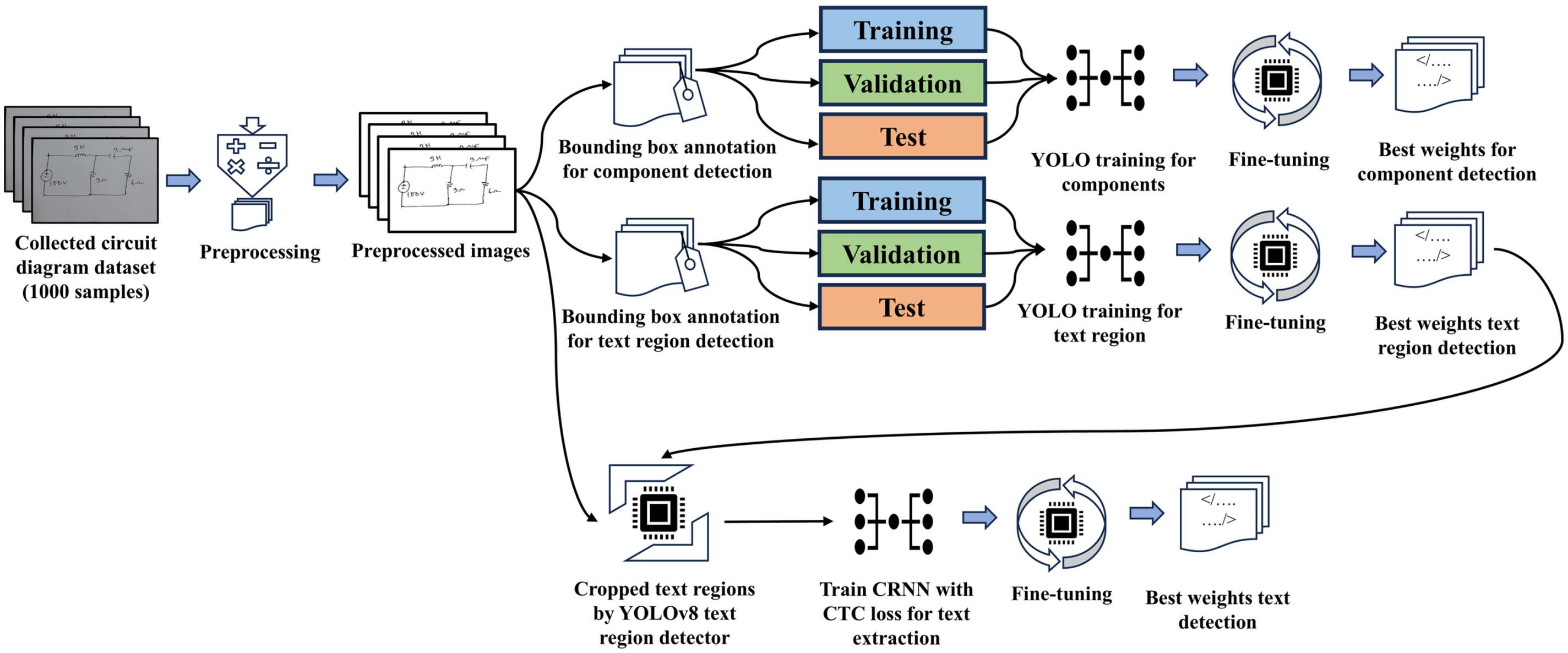

**Figure 5** **Training workflow showing YOLOv8 models for component and text-region detection, and CRNN with CTC for OCR, each fine-tuned to obtain the optimal model weights.** Full-size DOI: 10.7717/peerj-cs.4103/fig-5

## End-to-end pipeline overview

The overall circuit simulation system is depicted in Fig. 6. The process begins with an input image of a hand-drawn circuit. After being preprocessed in the same way as the proposed dataset images have been preprocessed, the image is passed into three branches. The first and second branches consist of deep learning models, whereas the third branch consists of classical techniques. Firstly, a YOLOv8 model has been used to detect the electrical elements of the circuit diagram. In the next branch, another YOLOv8 has been utilized to detect the text regions of handwritten labels of the component and crop the text automatically, and the text of these cropped images was extracted using the CRNN + CTC OCR model, which provides the component values (*e.g.*, 10 Ω, 3 μF, 5 V). Finally, the detected component and the extracted text have been mapped together.

In the third branch, all detected components and their corresponding text regions have been removed to keep only the wire structure. Then, the system identifies the junctions and endpoints using edge detection, skeletonization, and DBSCAN clustering to produce a set of detected nodes. These nodes are then mapped to the nearest component terminal.

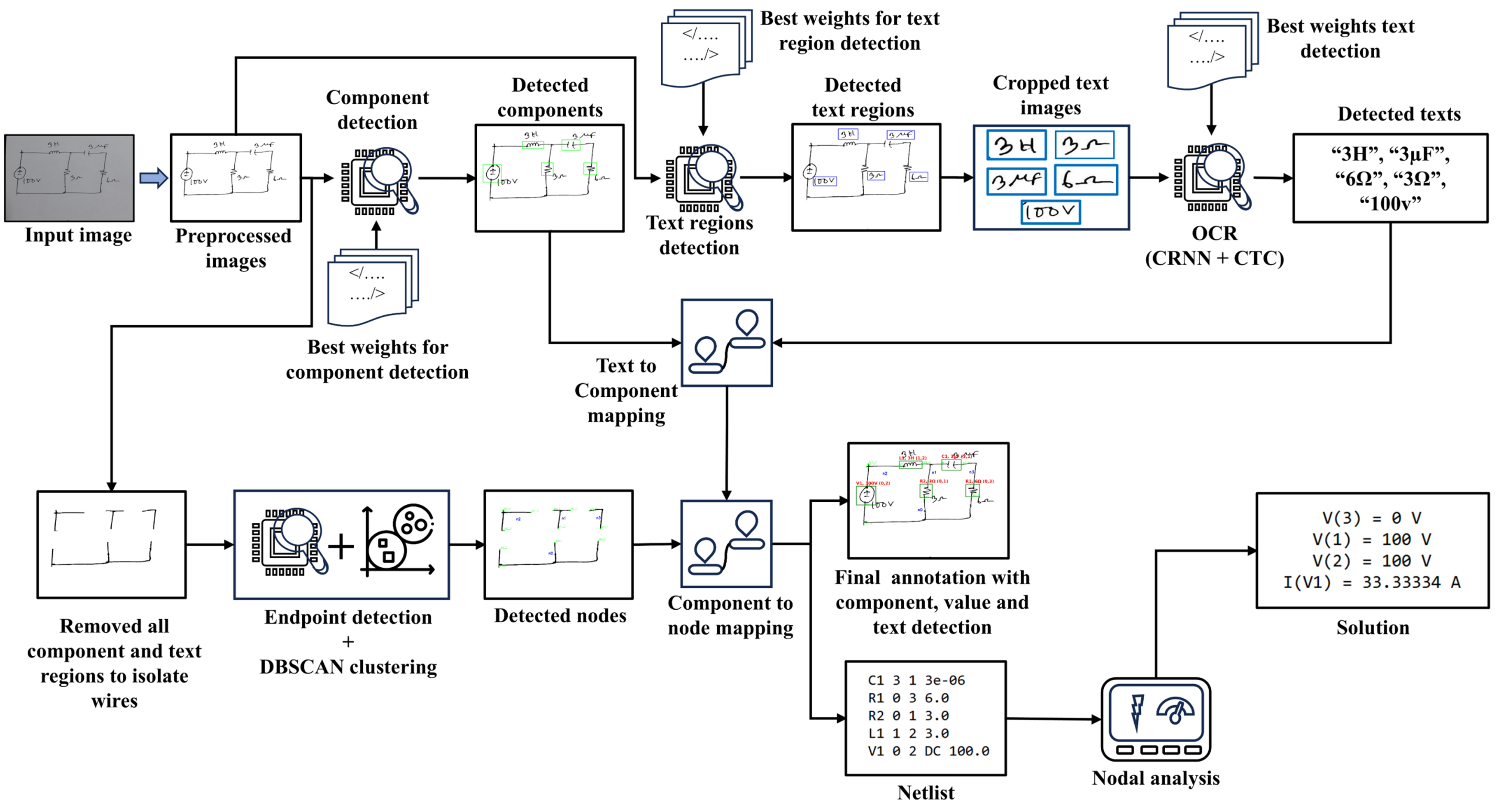

**Figure 6 End-to-end pipeline from the input image through component, text detection, OCR, node extraction, and component-to-node mapping to netlist generation and circuit solutions.** Full-size DOI: 10.7717/peerj-cs.4103/fig-6

Finally, the system combines the detected components, their values, and nodes to produce an annotated diagram and a SPICE-compatible netlist. The netlist is then solved using the DC nodal analysis method. The step-by-step process of building the simulation system will be described in the following subsections.

### *Component and text region detection*

Accurate component identification and text region localization are crucial, as errors in this phase will propagate through subsequent stages and eventually reduce the system's performance. Therefore, the YOLOv8 object detection framework was employed for both tasks because of its exceptional balance between its robustness to detect small objects in cluttered environments and low inference time (*Redmon et al., 2016*). YOLOv8 processes the entire image in one pass to predict bounding boxes, class probabilities, and confidence scores directly. Figure 7 illustrates the YOLOv8 architecture that comprises three modules: backbone, neck, and head.

The detection stage begins with an input of a processed 640 × 640 image to the YOLOv8 pipeline. The backbone module extracts multi-scale feature maps from the input image through alternating 3 × 3 convolution layers with CSP Bottleneck with 2 convolutions, faster (C2f) blocks and Spatial Pyramid Pooling-Fast (SPPF) layers. The Feature Pyramid Network (FPN)-Path Aggregation Network (PAN) neck fuses them and passes them to the anchor-free decoupled head that produces bounding boxes with class confidences. Then, non-maximum suppression (NMS) is applied to remove overlaps (Intersection over Union (IoU) threshold 0.5). Finally, the component detector model produce $B_c = \{(b_i,\ y_i)\}_{i=1}^{N_c}$, where each box $b_i = (x_{1i},\ y_{1i},\ x_{2i},\ y_{2i}) \in \mathbb{R}^4$ and the class label $y_i \in \boldsymbol{y_c} = \{R,\ L,\ C,\ V_{dc},\ V_{ac},\ I_{dc}\}$; and the text-region detector model produce $B_t = \left\{\left(b_j,\ y_j^{(t)}\right)\right\}_{j=1}^{N_t}$, where each box $b_j = \left(x_{1j},\ y_{1j},\ x_{2j},\ y_{2j}\right) \in \mathbb{R}^4$ and the class label $y_j^{(t)} \in \boldsymbol{y_t} = \{text\}$ (single class). Here, $N_c$ and $N_t$ is the total number of component and text-region detected box returned by the models for an image.

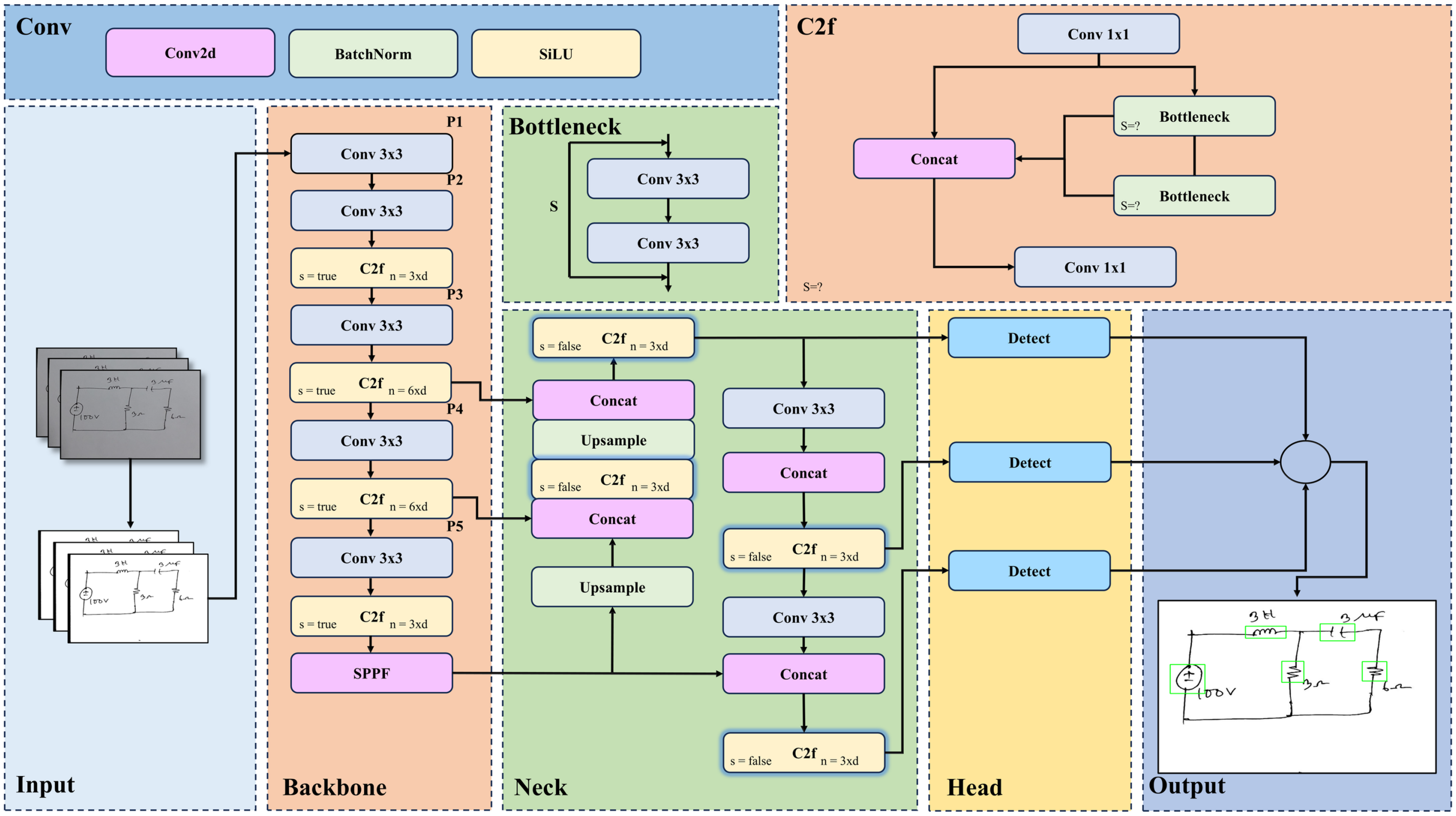

**Figure 7** **YOLOv8 architecture showing the backbone (feature extraction), neck (multiscale feature fusion), and head (anchor-free detection) used for component recognition.** Full-size DOI: 10.7717/peerj-cs.4103/fig-7

### Text extraction (CRNN + CTC)

Using the bounding box of the text-region detector YOLOv8 model, handwritten labels are cropped and then fed into an OCR engine for text recognition. The OCR is implemented using a CRNN combined with CTC loss as depicted in Fig. 8. The architecture is designed specifically for sequence-based text recognition without requiring character-level segmentation.

The process starts with an input image of 32 × 128 pixels that passes through three blocks of convolutional layers with batch normalization, ReLU activations, and max pooling to produce compact feature maps that capture the strokes and shapes of handwritten characters. These feature maps are then processed by BiLSTM layers, where the engine learns the temporal order of strokes. Finally, a dense layer with softmax activation and CTC loss is used. As illustrated in Fig. 8, the CTC decoder collapses repeats and blanks, then provides the final string (*e.g.*, "4 μF" from the sequence "- 4 4 - μ μ - F F -").

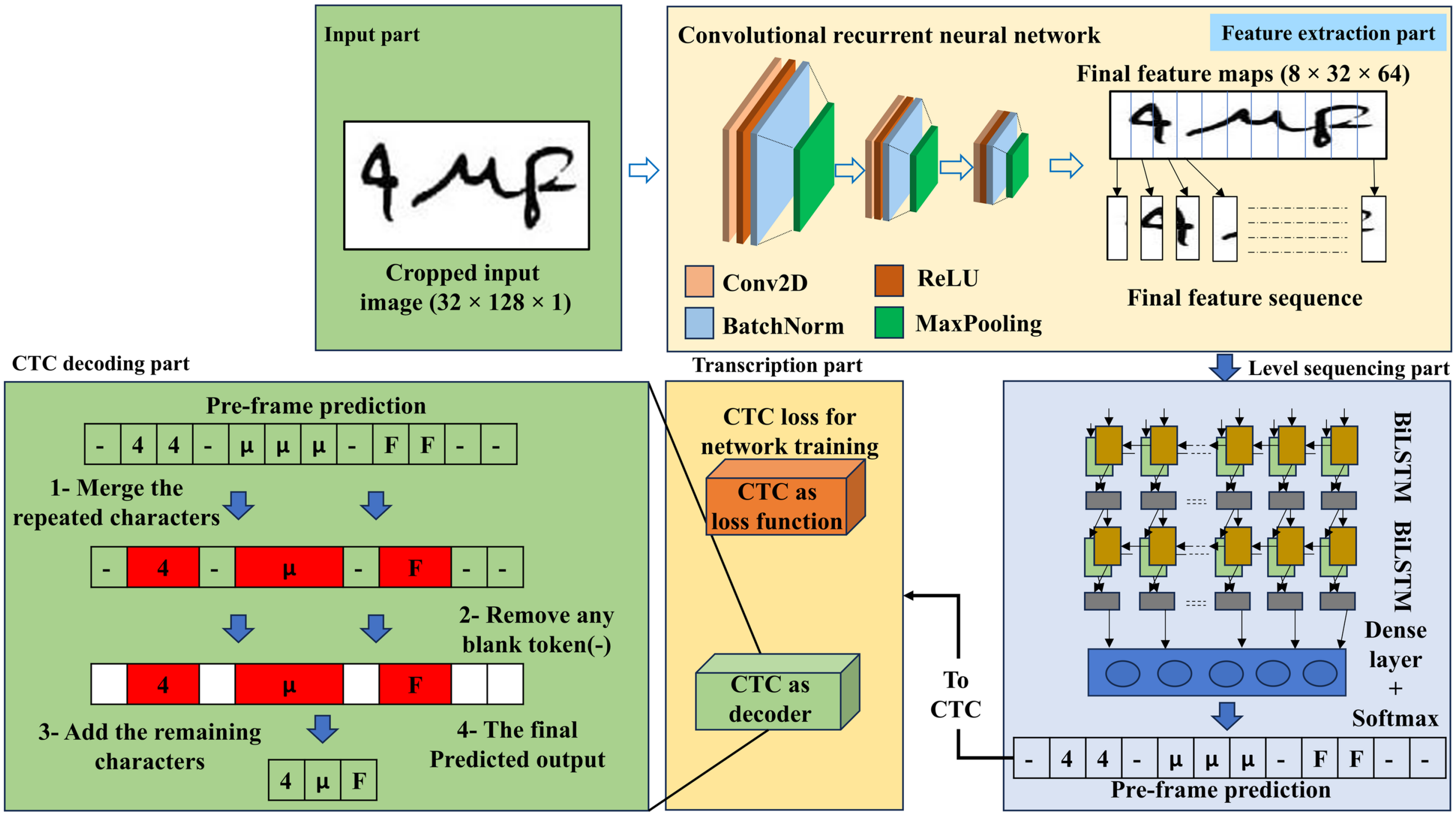

**Figure 8 Workflow of the CRNN with CTC for text recognition.** The convolutional layers extract features, the BiLSTM layers capture the sequence order, and the CTC decoder transcribes the final output. Full-size DOI: 10.7717/peerj-cs.4103/fig-8

### *Wire-node detection, text-components and component-node mapping*

After detecting all components $\{(b_i)\}_{i=1}^{N_c}$ and text-region $\{(b_j)\}_{j=1}^{N_t}$ boxes, they have been removed from the circuit diagram to form a masked binary image $I_{bin}$ consisting only of wires. Edges were then extracted from which all candidate endpoints $\mathcal{P}$ (degree = 1) and junctions $\mathcal{J}$ (degree ≥ 3) have been calculated through skeletonization. The candidate $\mathcal{P}$'s and $\mathcal{J}$'s are then spatially clustered with DBSCAN to remove jitter and fuse near-duplicate detections, yielding a stable node set $\mathcal{N} = \{n_k\}_{\{k=1\}}^{N_n}$. The lowest node of the circuit diagram is considered as ground ($n_0 = 0$) and closer nodes (within τ radius) are merged.

Each component box $b_i$, Its four side midpoints $m_i^L,\ m_i^R,\ m_i^T,\ m_i^B$ are calculated, and a nearest-neighbor search has been performed over $\mathcal{N}$. The two closest distinct nodes $(n_{i1},\ n_{i2})$ are then assigned to the component terminals. If there is only one node, then by keeping it, another dummy node (999) is assigned to the other terminal, preserving the ground. Finally, text boxes are linked to the nearest component centroids in a way that

**Algorithm 1** **Wire/node detection and component-node mapping.**

**Inputs:** $I_{bin}$ (masked binary image),
$B_c = \{(b_i, y_i)\}_{i=1}^{N_c}$ ,$B_t = \{(b_j, y_j^{(t)})\}_{j=1}^{N_t}$

**Outputs:** Node set $\mathcal{N} = \{n_k\}_{\{k=1\}}^{N_n}$ (with ground $n_0$)
Component–node pairs $\{(b_i, n_{i1}, n_{i2})\}$,
Text→component links$\{(b_j, i^*)\}$

**Steps:**
1. $E \leftarrow \text{Canny}(I_{bin})$
2. $S \leftarrow \mathcal{S}(E)$
3. For each $p \in S$: $deg(p) \leftarrow \sum_{\{q \in \mathcal{N}_8(p)\}} \mathbb{1}[q \in S]$
4. $\mathcal{P} \leftarrow \{p: deg(p) = 1\}$ , $\mathcal{J} \leftarrow \{p: \deg(p) \geq 3\}$
5. $\mathcal{N} \leftarrow \mathcal{C}_{DBSCAN}(\mathcal{P} \cup \mathcal{J}; \varepsilon, \text{minPts})$
6. $n_0 \leftarrow \text{argmax}_{\{n \in \mathcal{N}\}} y(n)$; assign ID 0
7. Merge any $(n_k, n_\ell)$ with$||n_k - n_\ell||^2 < \tau$
8. For each component $b_i = (x_{1i}, y_{1i}, x_{2i}, y_{2i})$:
   - Compute side midpoints: $m_i^L = \left(x_{1i}, \frac{y_{1i}+y_{2i}}{2}\right)$, $m_i^R = \left(x_{2i}, \frac{y_{1i}+y_{2i}}{2}\right)$, $m_i^T = \left(\frac{x_{1i}+x_{2i}}{2}, y_{1i}\right)$, $m_i^B = \left(\frac{x_{1i}+x_{2i}}{2}, y_{2i}\right)$
   - Distances: $d_{i,s,k} = ||m_i^s - n_k||_2$ for $s \in \{L, R, T, B\}$, $k = 1, \ldots, N_n$
   - Assign two closest distinct nodes $(n_{i1}, n_{i2})$; if only one, set other to 999
9. For each text box $b_j$:
   - $c_j = \left(\frac{x_{1j}+x_{2j}}{2}, \frac{y_{1j}+y_{2j}}{2}\right)$, $g_i = \left(\frac{x_{1i}+x_{2i}}{2}, \frac{y_{1i}+y_{2i}}{2}\right)$,
   - $i^* = \text{argmin}_{1 \leq i \leq N_c} ||c_j - g_i||_2$
10. Return $\mathcal{N}, \{(b_i, n_{i1}, n_{i2})\}_{i=1}^{N_c}, \{(b_j, i^*)\}_{j=1}^{N_t}$

**Table 2** **Corresponding output and values of Algorithm 1.**

| Component | Detected class | OCR value | Connected wires | Mapped nodes |
|---|---|---|---|---|
| C1 | Capacitor | 3 μF | Wire 1, Wire 3 | $n_1$–$n_3$ |
| R1 | Resistor | 6 Ω | Wire 3, Wire 0 | $n_3$–$n_0$ |
| R2 | Resistor | 3 Ω | Wire 1, Wire 0 | $n_1$–$n_0$ |
| L1 | Inductor | 3 H | Wire 2, Wire 1 | $n_2$–$n_1$ |
| V1 | DC voltage source | 100 V | Wire 2, Wire 0 | $n_2$–$n_0$ |

OCR values are attached to the correct instances. Algorithm 1 summarizes the step-by-step process, and Table 2, along with Fig. 9, presents the output of this unified algorithm.

### *Final annotation, netlist generation, and nodal analysis*

Based on the data of the previous sub-section, a visual overlay and simulation-ready description is produced. Each component $i$ gets its value $v_i$ by parsing $\hat{s}_j$ and unit mapping. The detector classes are then mapped to a SPICE prefix $\pi(y_i) \in \{R, L, C, V, I\}$. Each element is formatted with its terminals $(n_{i1}.\ n_{i2})$ and any terminal of 999 is omitted from the simulation but retained in the overlay. The final compact label $[\pi(y_i)k,\ \hat{s}_j$ or "No Value", $(n_{i1},\ n_{i2})]$ is attached to each $b_i$. Each label in the diagram is analogous to each SPICE line, and by concatenating all the lines, header, and footer, the final netlist is produced. In the end, modified nodal analysis has been applied to the generated netlist to generate the final solution. The overall process is described in Algorithm 2, and its corresponding response is depicted in Fig. 10.

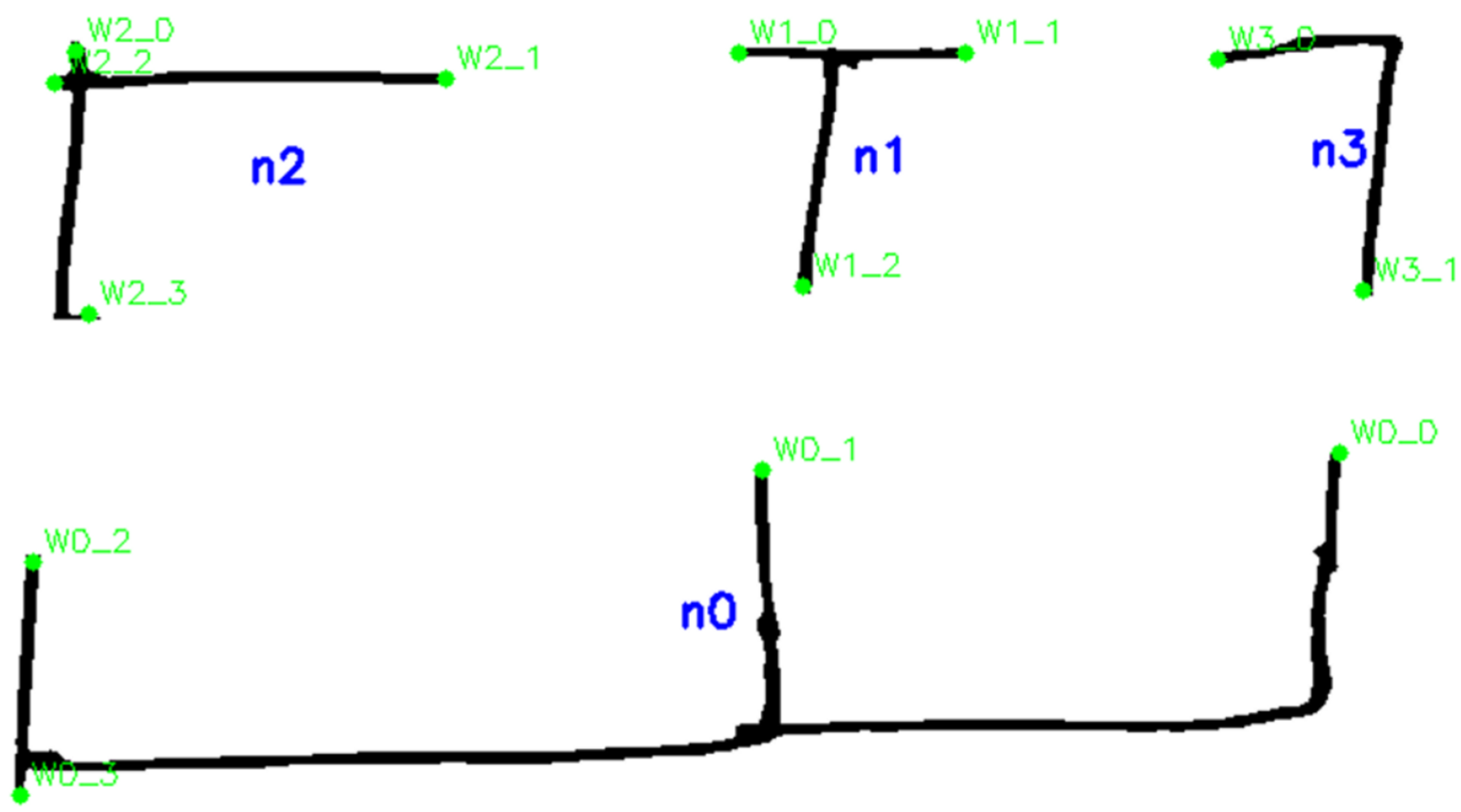

**Figure 9** Detected node clusters and wire endpoints used for component-to-node mapping.
Full-size DOI: 10.7717/peerj-cs.4103/fig-9

**Algorithm 2** Final annotation, netlist generation, and nodal analysis.

**Inputs:** $\mathcal{N} = \{n_k\}_{\{k=1\}}^{N_n}$ (ground $n_0 = 0$); terminals$(n_{i1}, n_{i2})$
$B_c = \{(b_i, y_i)\}_{i=1}^{N_c}$; Text→component links $\{(b_j, i^*)\}$,
OCR strings $\{\hat{s}_j\}$; unit map $m{:}u \to \mathbb{R}^+$;
class map $\pi$: $y_c \to \{R, L, C, V, I\}$; nominal $f_0$

**Outputs:** $\{lab_i\}$, Netlist $\mathcal{L}$, DC$\{V(n)\}$, $\{I(V_k)\}$

**Steps:**
1. For $i = 1, \ldots, N_c$
   Value: $j \leftarrow i^*$; parse $\hat{s}_j \to (a_i, v_i)$; $v_i \leftarrow \begin{cases} a_i\ m(u_i), & ok \\ 1, & else \end{cases}$
   Prefix: $p_i \leftarrow \pi(y_i)$
   $lab_i = [p_i k, \hat{s}_j$ or "No Value", $(n_{i1}, n_{i2})]$
2. Element line:
   $$\ell_i = \begin{cases} V_i\ n_{i1}\ n_{i2}\ DC\ v_i, & y_i = V_{dc}, \\ V_i\ n_{i1}\ n_{i2}\ AC\ v_i f_0, & y_i = V_{ac}, \\ I_i\ n_{i1}\ n_{i2}\ v_{i,} & y_i = I_{dc}, \\ p_i i\ n_{i1}\ n_{i2}\ v_{i,} & y_i \in \{R, L, C\}. \end{cases}$$
   If $n_{i1} = 999$ or $n_{i2} = 999 \to$ tag incomplete (omit)
3. Netlist assembly
   $\mathcal{L} \leftarrow$ [* Autogenerated; .ground 0; $\ell_i$; ...; $\ell_{N_c}$; .tran 1 m 1; .end]
4. DC MNA
   - Nodes $v = \{n \neq 0\}$ (size $N$); voltage sources list size M; Unknown $x \in \mathbb{R}^{\mathbb{N}+\mathbb{M}}$
   - Initialize $G \in \mathbb{R}^{(N+M)\times(N+M)} = 0, I \in \mathbb{R}^{(N+M)} = 0$
   - Stamp (per element):
     - Resistor $R$ $a \leftrightarrow b$: $g = \frac{1}{R}$, add $\begin{bmatrix} g & -g \\ -g & g \end{bmatrix}$ to $(a, b)$ block
     - Current $I_0$: $a \to b$: $I_a - = I_0, I_b + = I_0$
     - Inductor (DC): short, use $g_{short} \gg 1$ as above
     - Capacitor (DC): open, no stamp.
     - Voltage $V_0$: $a \leftrightarrow b$: add row/col $k = N+$ index;
       $G_{a,k} + = 1, G_{b,k} + = 1, G_{k,a} + = 1, G_{k,b} + = 1, I_k = V_0$.
   - Solve $G_x = I$; report $V(n)$ from the first $N$ entires;
     $I(V_k)$ from the last $M$.

5. Return $(\{lab_i\})$, $\mathcal{L}$, $\{V(n)\}$,$\{I(V_k)\}$)

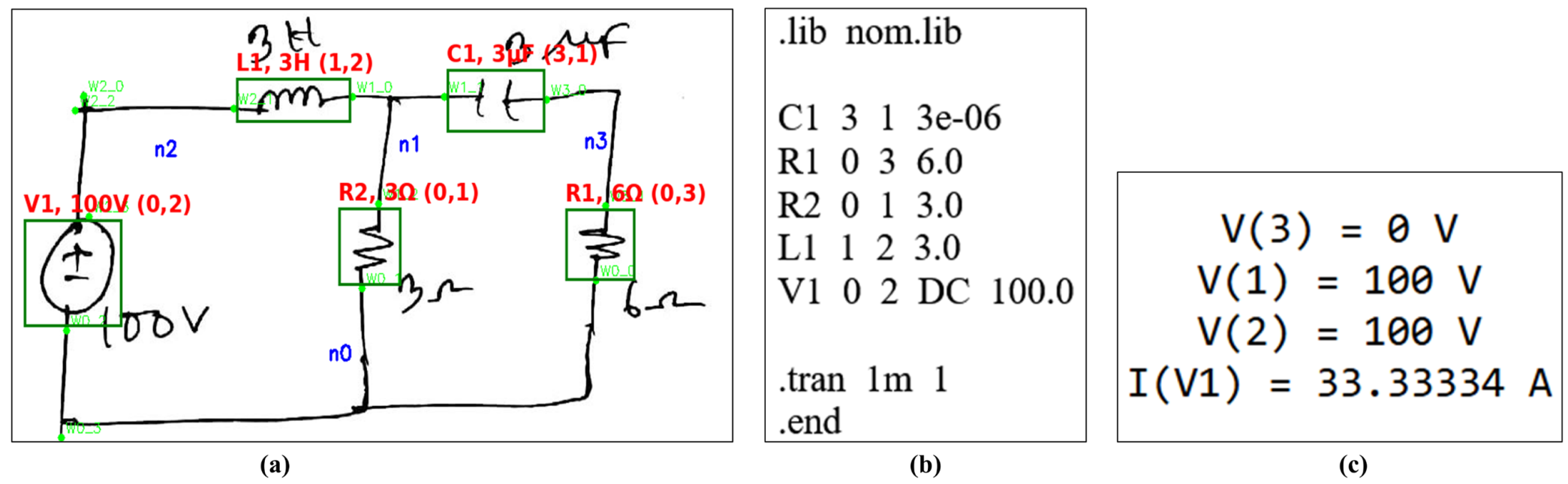

**Figure 10** **Outputs of Algorithm 2 (A) final annotated circuit, (B) generated netlist and (C) solution of the circuits *via* nodal analysis.** Full-size DOI: 10.7717/peerj-cs.4103/fig-10

# RESULT ANALYSIS

This section provides the empirical evaluation of the proposed multi-stage pipeline for handwritten circuit simulation. The analysis is segmented into the selection of an optimal YOLOV8 model for component detection, text region localization, and performance analysis of the CRNN-based text recognition module. The experiments were conducted on the proposed custom dataset of 1,000 hand-drawn diagrams, utilizing several performance metrics. Ablation studies technically varied hyperparameters to investigate their impacts, trade-offs in model scaling, and generalization. Finally, the holistic efficacy of end-to-end circuit interpretation is demonstrated using several examples.

## Component detection results

Component detection has been assessed across six different classes using three YOLOv8 variants- nano, small, and medium. The detailed results are presented in Tables 3–5. The choice of optimizer, learning rate, and freezing layers was highly consequential for all the models' performance. RMSprop catastrophically dropped the model's performance, whereas other optimizers nearly yielded the same results. The best performance was obtained with learning rates between 0.005 and 0.01, while extremely small or large learning rates reduced performance. A similar pattern has been observed for freezing layers. As the number of freezing layers increased, all the models' performance was reduced. Hyperparameters like the number of epochs and image size had little significance on model performance; however, they had a noticeable impact on training time. Freezing layer is the only hyperparameter that influenced both training time and model performance. The rest of the hyperparameters, such as weight decay, batch size, and dropout rate no notable impact on the models.

Based on the ablation study of Tables 3–5, the best trade-off hyperparameter was selected. The final hypermeter settings for nano, small, and medium models are described in Table 6. After fine-tuning YOLOv8 variants with these settings, they were compared against each other. Table 7 presents the class-wise analysis of the models. However, due to the limited size of the evaluation dataset, the reported differences between YOLO variants

**Table 3** Ablation study of YOLOv8-nano for component detection.

**Case exploration 1: Modifying epoch**

| Ex. no. | Epoch | Training time | Precision | Recall | mAP@0.5 | mAP@0.5:0.95 | Discovering |
|---|---|---|---|---|---|---|---|
| 1 | 16 | 196.92 | 0.985257 | 0.984612 | 0.987261 | 0.668119 | Balanced baseline |
| 2 | 32 | 370.93 | 0.981088 | 0.989242 | 0.9877 | 0.682525 | Recall improved |
| 3 | 64 | 714.74 | 0.985945 | 0.986906 | 0.984838 | 0.689188 | Best mAP@0.5:0.95 |

**Case exploration 2: Modifying weight decay**

| Ex. no. | Weight decay | Training time | Precision | Recall | mAP@0.5 | mAP@0.5:0.95 | Discovering |
|---|---|---|---|---|---|---|---|
| 1 | 0 | 369.78 | 0.986227 | 0.988836 | 0.98616 | 0.678267 | Lower mAP@0.5:0.95 |
| 2 | 0.0001 | 371.61 | 0.985529 | 0.986107 | 0.986596 | 0.678419 | Balanced performance |
| 3 | 0.0005 | 370.07 | 0.981088 | 0.989242 | 0.9877 | 0.682525 | Best mAP@0.5:0.95 |
| 4 | 0.001 | 368.9 | 0.986666 | 0.986721 | 0.988118 | 0.680819 | Balanced performance |
| 5 | 0.0005 | 368.17 | 0.988092 | 0.986854 | 0.987962 | 0.678105 | Highest precision |

**Case exploration 3: Modifying batch size**

| Ex. no. | Batch size | Training time | Precision | Recall | mAP@0.5 | mAP@0.5:0.95 | Discovering |
|---|---|---|---|---|---|---|---|
| 1 | 16 | 423.51 | 0.984903 | 0.989191 | 0.987963 | 0.685118 | Best Recall, higher mAP |
| 2 | 32 | 366.51 | 0.981088 | 0.989242 | 0.9877 | 0.682525 | Balanced baseline |
| 3 | 64 | 367.07 | 0.983601 | 0.988387 | 0.988297 | 0.683367 | Highest mAP@0.5 |

**Case exploration 4: Modifying dropout rate**

| Ex. no. | Dropout rate | Training time | Precision | Recall | mAP@0.5 | mAP@0.5:0.95 | Discovering |
|---|---|---|---|---|---|---|---|
| 1 | 0 | 367.14 | 0.981088 | 0.989242 | 0.9877 | 0.682525 | Baseline |
| 2 | 0.1 | 367.18 | 0.981088 | 0.989242 | 0.9877 | 0.682525 | No change |
| 3 | 0.2 | 372.42 | 0.981088 | 0.989242 | 0.9877 | 0.682525 | No change |
| 4 | 0.3 | 383.74 | 0.981088 | 0.989242 | 0.9877 | 0.682525 | No change |
| 5 | 0.5 | 378.15 | 0.981088 | 0.989242 | 0.9877 | 0.682525 | No change |

**Case exploration 5: Modifying image size**

| Ex. no. | Image size | Training time | Precision | Recall | mAP@0.5 | mAP@0.5:0.95 | Discovering |
|---|---|---|---|---|---|---|---|
| 1 | 320 | 180.4 | 0.982313 | 0.986128 | 0.987097 | 0.66823 | Lower mAP@0.5:0.95 |
| 2 | 512 | 284.4 | 0.990532 | 0.986219 | 0.985877 | 0.669533 | Highest Precision |
| 3 | 640 | 377.19 | 0.981088 | 0.989242 | 0.9877 | 0.682525 | Best mAP@0.5:0.95 |

**Case exploration 6: Modifying optimizer**

| Ex. no. | Optimizer | Training time | Precision | Recall | mAP@0.5 | mAP@0.5:0.95 | Discovering |
|---|---|---|---|---|---|---|---|
| 1 | SGD | 362.24 | 0.981088 | 0.989242 | 0.9877 | 0.682525 | Balanced baseline |
| 2 | Adam | 369.41 | 0.984315 | 0.984981 | 0.987161 | 0.686147 | Improved stability |
| 3 | AdamW | 364.1 | 0.984601 | 0.983345 | 0.987049 | 0.685194 | Comparable results |
| 4 | RMSProp | 369.6 | 0.609359 | 0.666548 | 0.643663 | 0.256353 | Accuracy dropped |
| 5 | Adamax | 361.97 | 0.987643 | 0.985954 | 0.988209 | 0.690002 | Best overall |

**Case exploration 7: Modifying learning rate**

| Ex. no. | Learning rate | Training time | Precision | Recall | mAP@0.5 | mAP@0.5:0.95 | Discovering |
|---|---|---|---|---|---|---|---|
| 1 | 0.001 | 364.24 | 0.987371 | 0.985933 | 0.986992 | 0.679253 | Lower mAP@0.5:0.95 |
| 2 | 0.005 | 363.5 | 0.98411 | 0.987906 | 0.987858 | 0.691069 | Best mAP@0.5:0.95 |

**Table 3** (continued)

**Case exploration 7: Modifying learning rate**

| Ex. no. | Learning rate | Training time | Precision | Recall | mAP@0.5 | mAP@0.5:0.95 | Discovering |
|---|---|---|---|---|---|---|---|
| 3 | 0.01 | 365.09 | 0.987643 | 0.985954 | 0.988209 | 0.690002 | Best trade-off |
| 4 | 0.05 | 404.92 | 0.98733 | 0.983136 | 0.987154 | 0.679864 | Lower mAP@0.5:0.95 |
| 5 | 0.1 | 371.4 | 0.957043 | 0.97203 | 0.982569 | 0.660852 | Accuracy dropped |

**Case exploration 8: Freezing layers**

| Ex. no. | Freezing layers | Training time | Precision | Recall | mAP@0.5 | mAP@0.5:0.95 | Discovering |
|---|---|---|---|---|---|---|---|
| 1 | 0 | 377.07 | 0.98411 | 0.987906 | 0.987858 | 0.691069 | Best mAP@0.5:0.95 |
| 2 | 10 | 315.79 | 0.981012 | 0.98533 | 0.987636 | 0.681068 | Accuracy stable |
| 3 | 15 | 298.7 | 0.978978 | 0.978054 | 0.985193 | 0.671502 | Accuracy dropped |
| 4 | 22 | 291.13 | 0.931456 | 0.94868 | 0.972238 | 0.625439 | Accuracy dropped |

**Case exploration 9: Activation function**

| Ex. no. | Activation function | Training time | Precision | Recall | mAP@0.5 | mAP@0.5:0.95 | Discovering |
|---|---|---|---|---|---|---|---|
| 1 | ReLU | 401.46 | 0.975844 | 0.971732 | 0.984732 | 0.660244 | Accuracy dropped |
| 2 | Mish | 386.61 | 0.97957 | 0.971312 | 0.985431 | 0.666438 | Stable performance |
| 3 | LeakyReLU | 384.44 | 0.971935 | 0.973006 | 0.98368 | 0.664078 | Accuracy dropped |
| 4 | SiLU | 384.33 | 0.97642 | 0.98137 | 0.98462 | 0.666359 | Accuracy dropped |

**Table 4** Ablation study of YOLOv8-small for component detection.

**Case exploration 1: Modifying epoch**

| Ex. no. | Epoch | Training time | Precision | Recall | mAP@0.5 | mAP@0.5:0.95 | Discovering |
|---|---|---|---|---|---|---|---|
| 1 | 16 | 331.73 | 0.982379 | 0.986437 | 0.986795 | 0.683501 | Lower mAP@0.5:0.95 |
| 2 | 32 | 621.05 | 0.988021 | 0.983761 | 0.987769 | 0.690482 | Best performance |
| 3 | 64 | 1,218.83 | 0.987321 | 0.987326 | 0.988157 | 0.685919 | Higher training time |

**Case exploration 2: Modifying weight decay**

| Ex. no. | Weight decay | Training time | Precision | Recall | mAP@0.5 | mAP@0.5:0.95 | Discovering |
|---|---|---|---|---|---|---|---|
| 1 | 0 | 615.26 | 0.986476 | 0.986628 | 0.988854 | 0.687352 | Lower mAP@0.5:0.95 |
| 2 | 0.0001 | 619.79 | 0.985917 | 0.988823 | 0.988066 | 0.690485 | Best performance |
| 3 | 0.0005 | 622.55 | 0.988021 | 0.983761 | 0.987769 | 0.690482 | Balanced performance |
| 4 | 0.001 | 616.64 | 0.984984 | 0.987742 | 0.987569 | 0.69087 | Balanced performance |
| 5 | 0.005 | 617.03 | 0.98707 | 0.986474 | 0.987501 | 0.685681 | Slightly lower mAP |

**Case exploration 3: Modifying batch size**

| Ex. no. | Batch size | Training time | Precision | Recall | mAP@0.5 | mAP@0.5:0.95 | Discovering |
|---|---|---|---|---|---|---|---|
| 1 | 16 | 697.55 | 0.987834 | 0.985362 | 0.98836 | 0.684968 | Slightly lower mAP |
| 2 | 32 | 616.49 | 0.984984 | 0.987742 | 0.987569 | 0.69087 | Balanced baseline |
| 3 | 64 | 614.82 | 0.985493 | 0.987132 | 0.986153 | 0.692973 | Highest mAP@0.5:0.95 |

**Case exploration 4: Modifying dropout rate**

| Ex. no. | Dropout rate | Training time | Precision | Recall | mAP@0.5 | mAP@0.5:0.95 | Discovering |
|---|---|---|---|---|---|---|---|
| 1 | 0 | 624.12 | 0.984984 | 0.987742 | 0.987569 | 0.69087 | Baseline |
| 2 | 0.1 | 623.69 | 0.984984 | 0.987742 | 0.987569 | 0.69087 | No change |

(Continued)

Table 4 (continued)

| Case exploration 4: Modifying dropout rate | | | | | | | |
|---|---|---|---|---|---|---|---|
| **Ex. no.** | **Dropout rate** | **Training time** | **Precision** | **Recall** | **mAP@0.5** | **mAP@0.5:0.95** | **Discovering** |
| 3 | 0.2 | 626.31 | 0.984984 | 0.987742 | 0.987569 | 0.69087 | No change |
| 4 | 0.3 | 634.32 | 0.984984 | 0.987742 | 0.987569 | 0.69087 | No change |
| 5 | 0.5 | 626.37 | 0.984984 | 0.987742 | 0.987569 | 0.69087 | No change |
| **Case exploration 5: Modifying image size** | | | | | | | |
| **Ex. no.** | **Image size** | **Training time** | **Precision** | **Recall** | **mAP@0.5** | **mAP@0.5:0.95** | **Discovering** |
| 1 | 320 | 258.29 | 0.987315 | 0.984069 | 0.986354 | 0.687416 | Lower mAP@0.5:0.95 |
| 2 | 512 | 436.93 | 0.984842 | 0.986426 | 0.986981 | 0.692963 | Highest mAP |
| 3 | 640 | 611.48 | 0.984984 | 0.987742 | 0.987569 | 0.690874 | Best Recall & mAP |
| **Case exploration 6: Modifying optimizer** | | | | | | | |
| **Ex. no.** | **Optimizer** | **Training time** | **Precision** | **Recall** | **mAP@0.5** | **mAP@0.5:0.95** | **Discovering** |
| 1 | SGD | 589.41 | 0.984984 | 0.987742 | 0.987569 | 0.69087 | Balanced baseline |
| 2 | Adam | 589.07 | 0.987328 | 0.985335 | 0.987494 | 0.683184 | Improved stability |
| 3 | AdamW | 599.74 | 0.990259 | 0.986942 | 0.987795 | 0.689997 | Highest precision |
| 4 | RMSProp | 606.82 | 0.958786 | 0.956359 | 0.979154 | 0.653564 | Accuracy dropped |
| 5 | Adamax | 585.18 | 0.987416 | 0.989719 | 0.988994 | 0.691606 | Best overall |
| **Case exploration 7: Modifying learning rate** | | | | | | | |
| **Ex. no.** | **Learning rate** | **Training time** | **Precision** | **Recall** | **mAP@0.5** | **mAP@0.5:0.95** | **Discovering** |
| 1 | 0.001 | 616.96 | 0.985751 | 0.988611 | 0.98823 | 0.69493 | Best performance |
| 2 | 0.005 | 603.18 | 0.981352 | 0.989356 | 0.988795 | 0.691391 | Best performance |
| 3 | 0.01 | 623.02 | 0.990259 | 0.986942 | 0.987795 | 0.689997 | Best precision |
| 4 | 0.05 | 620.63 | 0.967707 | 0.964778 | 0.981485 | 0.659103 | Lower mAP@0.5:0.95 |
| 5 | 0.1 | 600.37 | 0.964743 | 0.94066 | 0.978373 | 0.633693 | Accuracy dropped |
| **Case exploration 8: Freezing layers** | | | | | | | |
| **Ex. no.** | **Freezing layers** | **Training time** | **Precision** | **Recall** | **mAP@0.5** | **mAP@0.5:0.95** | **Discovering** |
| 1 | 0 | 600.15 | 0.985751 | 0.988611 | 0.98823 | 0.69493 | Best performance |
| 2 | 10 | 397.28 | 0.987146 | 0.987632 | 0.989474 | 0.685735 | Accuracy stable |
| 3 | 15 | 386.31 | 0.988566 | 0.985344 | 0.988271 | 0.683634 | Accuracy dropped |
| 4 | 22 | 354.95 | 0.970979 | 0.975202 | 0.985015 | 0.636967 | Accuracy dropped |
| **Case exploration 9: Activation function** | | | | | | | |
| **Ex. no.** | **Activation function** | **Training time** | **Precision** | **Recall** | **mAP@0.5** | **mAP@0.5:0.95** | **Discovering** |
| 1 | ReLU | 586.5 | 0.985076 | 0.979921 | 0.985547 | 0.676133 | Accuracy dropped |
| 2 | Mish | 587.24 | 0.980769 | 0.982609 | 0.986435 | 0.673206 | Stable performance |
| 3 | LeakyReLU | 605.79 | 0.972495 | 0.977893 | 0.98684 | 0.678608 | Accuracy dropped |
| 4 | SiLU | 595.93 | 0.986167 | 0.98425 | 0.987227 | 0.68015 | Stable performance |

in Table 7 should be interpreted as indicative trends rather than statistically significant differences. Whereas Fig. 11 presents the parameter, training time, and accuracy-based comparison among the three variants of the YOLOv8 model.

**Table 5** Ablation study of YOLOv8-medium for component detection.

**Case exploration 1: Modifying epoch**

| Ex. no. | Epoch | Training time | Precision | Recall | mAP@0.5 | mAP@0.5:0.95 | Discovering |
|---|---|---|---|---|---|---|---|
| 1 | 16 | 613.52 | 0.989062 | 0.988428 | 0.986567 | 0.684767 | Lower mAP@0.5:0.95 |
| 2 | 32 | 1,172.43 | 0.989098 | 0.985632 | 0.988001 | 0.69146 | Best performance |
| 3 | 64 | 2,317.44 | 0.988015 | 0.989858 | 0.987396 | 0.687464 | Higher training time |

**Case exploration 2: Modifying weight decay**

| Ex. no. | Weight decay | Training time | Precision | Recall | mAP@0.5 | mAP@0.5:0.95 | Discovering |
|---|---|---|---|---|---|---|---|
| 1 | 0 | 1,171.26 | 0.991738 | 0.989974 | 0.989483 | 0.695146 | Lower mAP@0.5 |
| 2 | 0.0001 | 1,153.54 | 0.989941 | 0.988042 | 0.986847 | 0.697338 | Best performance |
| 3 | 0.0005 | 1,155.44 | 0.989098 | 0.985632 | 0.988001 | 0.69146 | Balanced performance |
| 4 | 0.001 | 1,155.79 | 0.991306 | 0.990508 | 0.987335 | 0.69197 | Balanced performance |
| 5 | 0.005 | 1,155.92 | 0.991672 | 0.989429 | 0.987744 | 0.691011 | Slightly lower Recall |

**Case exploration 3: Modifying batch size**

| Ex. no. | Batch size | Training time | Precision | Recall | mAP@0.5 | mAP@0.5:0.95 | Discovering |
|---|---|---|---|---|---|---|---|
| 1 | 16 | 1,267.75 | 0.989809 | 0.989471 | 0.987395 | 0.686635 | Slightly lower mAP |
| 2 | 32 | 1,153.51 | 0.991306 | 0.990505 | 0.987318 | 0.691921 | Balanced baseline |
| 3 | 64 | Error | Error | Error | Error | Error | Memory error |

**Case exploration 4: Modifying dropout rate**

| Ex. no. | Dropout rate | Training time | Precision | Recall | mAP@0.5 | mAP@0.5:0.95 | Discovering |
|---|---|---|---|---|---|---|---|
| 1 | 0 | 1,176.32 | 0.991306 | 0.990508 | 0.987335 | 0.69197 | Baseline |
| 2 | 0.1 | 1,161.46 | 0.991306 | 0.990508 | 0.987335 | 0.69197 | No change |
| 3 | 0.2 | 1,161.13 | 0.991306 | 0.990508 | 0.987335 | 0.69197 | No change |
| 4 | 0.3 | 1,160.67 | 0.991306 | 0.990508 | 0.987335 | 0.69197 | No change |
| 5 | 0.5 | 1,164.58 | 0.991306 | 0.990508 | 0.987335 | 0.69197 | No change |

**Case exploration 5: Modifying image size**

| Ex. no. | Image size | Training Time | Precision | Recall | mAP@0.5 | mAP@0.5:0.95 | Discovering |
|---|---|---|---|---|---|---|---|
| 1 | 320 | 366.28 | 0.99103 | 0.987418 | 0.987867 | 0.693141 | Lower mAP@0.5:0.95 |
| 2 | 512 | 944.04 | 0.98878 | 0.987842 | 0.988072 | 0.692527 | Highest mAP |
| 3 | 640 | 1,156.88 | 0.991306 | 0.990508 | 0.987335 | 0.69197 | Best Recall & mAP |

**Case exploration 6: Modifying optimizer**

| Ex. no. | Optimizer | Training time | Precision | Recall | mAP@0.5 | mAP@0.5:0.95 | Discovering |
|---|---|---|---|---|---|---|---|
| 1 | SGD | 1,161.23 | 0.991306 | 0.990508 | 0.987335 | 0.69197 | Best performance |
| 2 | Adam | 1,172.61 | 0.986126 | 0.986969 | 0.987008 | 0.681319 | Slightly lower mAP |
| 3 | AdamW | 1,168.31 | 0.986241 | 0.985495 | 0.988257 | 0.687184 | Slightly lower mAP |
| 4 | RMSProp | 1,219.09 | 0.284918 | 0.290732 | 0.277995 | 0.159001 | Accuracy dropped |
| 5 | Adamax | 1,171.89 | 0.988617 | 0.987192 | 0.988335 | 0.691345 | Best stability |

**Case exploration 7: Modifying learning rate**

| Ex. no. | Learning rate | Training time | Precision | Recall | mAP@0.5 | mAP@0.5:0.95 | Discovering |
|---|---|---|---|---|---|---|---|
| 1 | 0.001 | 1,157.34 | 0.990106 | 0.99038 | 0.987504 | 0.696559 | Best performance |
| 2 | 0.005 | 1,157.42 | 0.991307 | 0.990505 | 0.987316 | 0.691921 | Best performance |

(Continued)

Table 5 (continued)

| Case exploration 7: Modifying learning rate | | | | | | | |
|---|---|---|---|---|---|---|---|
| **Ex. no.** | **Learning rate** | **Training time** | **Precision** | **Recall** | **mAP@0.5** | **mAP@0.5:0.95** | **Discovering** |
| 3 | 0.01 | 1,158.61 | 0.987182 | 0.987035 | 0.988887 | 0.694668 | Lower precision |
| 4 | 0.05 | 1,168.65 | 0.984354 | 0.981298 | 0.988908 | 0.693403 | Lower Precision, Recall |
| **Case exploration 8: Freezing layers** | | | | | | | |
| **Ex. no.** | **Freezing layers** | **Training time** | **Precision** | **Recall** | **mAP@0.5** | **mAP@0.5:0.95** | **Discovering** |
| 1 | 0 | 1,172.43 | 0.989098 | 0.985632 | 0.988001 | 0.69146 | Best performance |
| 3 | 15 | 675.82 | 0.988366 | 0.986431 | 0.987107 | 0.689567 | Accuracy dropped |
| 4 | 22 | 504.21 | 0.978447 | 0.976879 | 0.98436 | 0.644947 | Accuracy dropped |
| **Case exploration 9: Activation function** | | | | | | | |
| **Ex. no.** | **Activation function** | **Training time** | **Precision** | **Recall** | **mAP@0.5** | **mAP@0.5:0.95** | **Discovering** |
| 1 | ReLU | 1,149.27 | 0.970877 | 0.981019 | 0.98492 | 0.662595 | Accuracy dropped |
| 2 | Mish | Error | Error | Error | Error | Error | Memory error |
| 3 | LeakyReLU | 1,151.51 | 0.979394 | 0.975902 | 0.984849 | 0.666269 | Accuracy dropped |
| 4 | SiLU | 1,155.79 | 0.991306 | 0.990508 | 0.987335 | 0.69197 | Stable performance |

Table 6 The best hyperparameter settings for the final nano, small, and medium models.

| Model name | Epoch | Weight decay | Batch size | Dropout | Image size | Optimizer | Learning rate | Freezing layers | Activation function |
|---|---|---|---|---|---|---|---|---|---|
| Nano | 64 | 0.0001 | 32 | 0.0 | 640 | Adamax | 0.001 | 0 | SiLU |
| Small | 32 | 0.001 | 64 | 0.0 | 640 | AdamW | 0.001 | 0 | SiLU |
| Medium | 32 | 0.0001 | 32 | 0.0 | 640 | SGD | 0.005 | 0 | SiLU |

Table 7 The class wise performance score of YOLOv8n/s/m using the hyperparameters of Table 6.

| YOLOv8n (Nano) | | | | |
|---|---|---|---|---|
| **Class** | **Precision** | **Recall** | **mAP@0.5** | **mAP@0.5:0.95** |
| DC voltage source | 0.989 | 0.972 | 0.981 | 0.757 |
| Resistor | 0.983 | 0.990 | 0.985 | 0.640 |
| Current source | 0.976 | 0.993 | 0.992 | 0.742 |
| Inductor | 0.993 | 0.989 | 0.992 | 0.643 |
| Capacitor | 0.993 | 0.989 | 0.985 | 0.604 |
| AC voltage source | 0.994 | 0.979 | 0.989 | 0.714 |
| All (mAP) | 0.988 | 0.985 | 0.986 | 0.683 |
| **YOLOv8s (Small)** | | | | |
| **Class** | **Precision** | **Recall** | **mAP@0.5** | **mAP@0.5:0.95** |
| DC voltage source | 0.988 | 0.979 | 0.985 | 0.781 |
| Resistor | 0.985 | 0.986 | 0.986 | 0.653 |
| Current source | 0.980 | 0.992 | 0.983 | 0.736 |
| Inductor | 0.990 | 0.989 | 0.987 | 0.660 |
| Capacitor | 0.992 | 0.989 | 0.990 | 0.619 |

**Table 7** (continued)

| YOLOv8s (Small) | | | | |
|---|---|---|---|---|
| **Class** | **Precision** | **Recall** | **mAP@0.5** | **mAP@0.5:0.95** |
| AC voltage source | 0.994 | 0.988 | 0.994 | 0.736 |
| All (mAP) | 0.988 | 0.987 | 0.987 | 0.697 |
| **YOLOv8m (Medium)** | | | | |
| **Class** | **Precision** | **Recall** | **mAP@0.5** | **mAP@0.5:0.95** |
| DC voltage source | 0.998 | 0.951 | 0.983 | 0.756 |
| Resistor | 0.995 | 0.994 | 0.987 | 0.652 |
| Current source | 0.980 | 0.988 | 0.986 | 0.742 |
| Inductor | 0.993 | 0.983 | 0.994 | 0.655 |
| Capacitor | 0.994 | 0.989 | 0.987 | 0.622 |
| AC voltage source | 0.992 | 0.994 | 0.995 | 0.727 |
| All (mAP) | 0.990 | 0.983 | 0.988 | 0.692 |

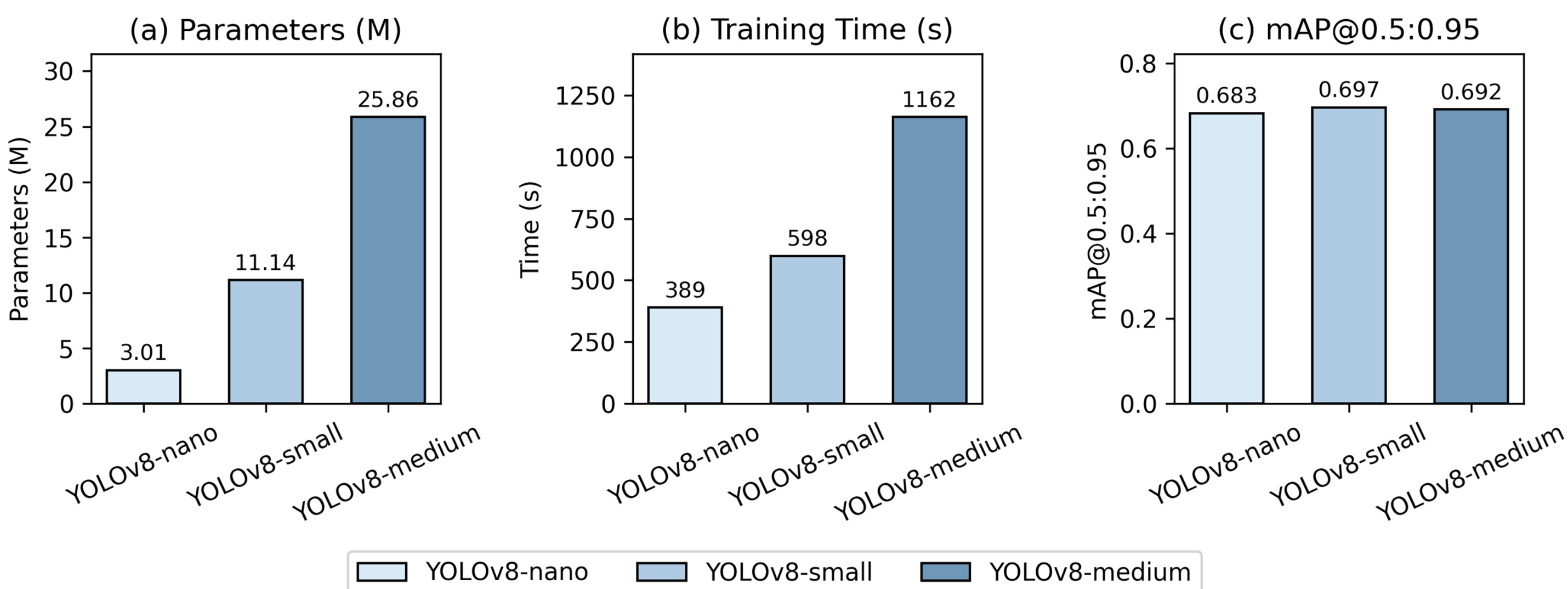

**Figure 11** **Comparison of YOLOv8-nano, YOLOv8-small, and YOLOv8-medium models in terms of (A) number of parameters, (B) training time, and (C) detection accuracy measured by mAP@0.5:0.95.** The trade-off between model complexity, training time, and detection performance is shown. Full-size DOI: 10.7717/peerj-cs.4103/fig-11

A critical finding is the point of diminishing return. As expected, YOLOv8-nano is the most lightweight (3.10M parameters) and served as the rapid model with the shortest training time (389 s). However, it yields the lowest performance (mean average precision (mAP)@0.5:0.95 of 0.683) when compared to the other two models. The Yolov8-small model (11.14M params) achieved the highest observed accuracy (mAP@0.5:0.95 of 0.697) in 580 s. On the other hand, the Yolov8-medium model, having 57% more parameters and requiring almost 2× training time, did not provide a meaningful improvement. This makes the Yolov8-small model the "sweet spot" offering the best balance between resource and performance. Hence, YOLOv8-small was selected for component detection.

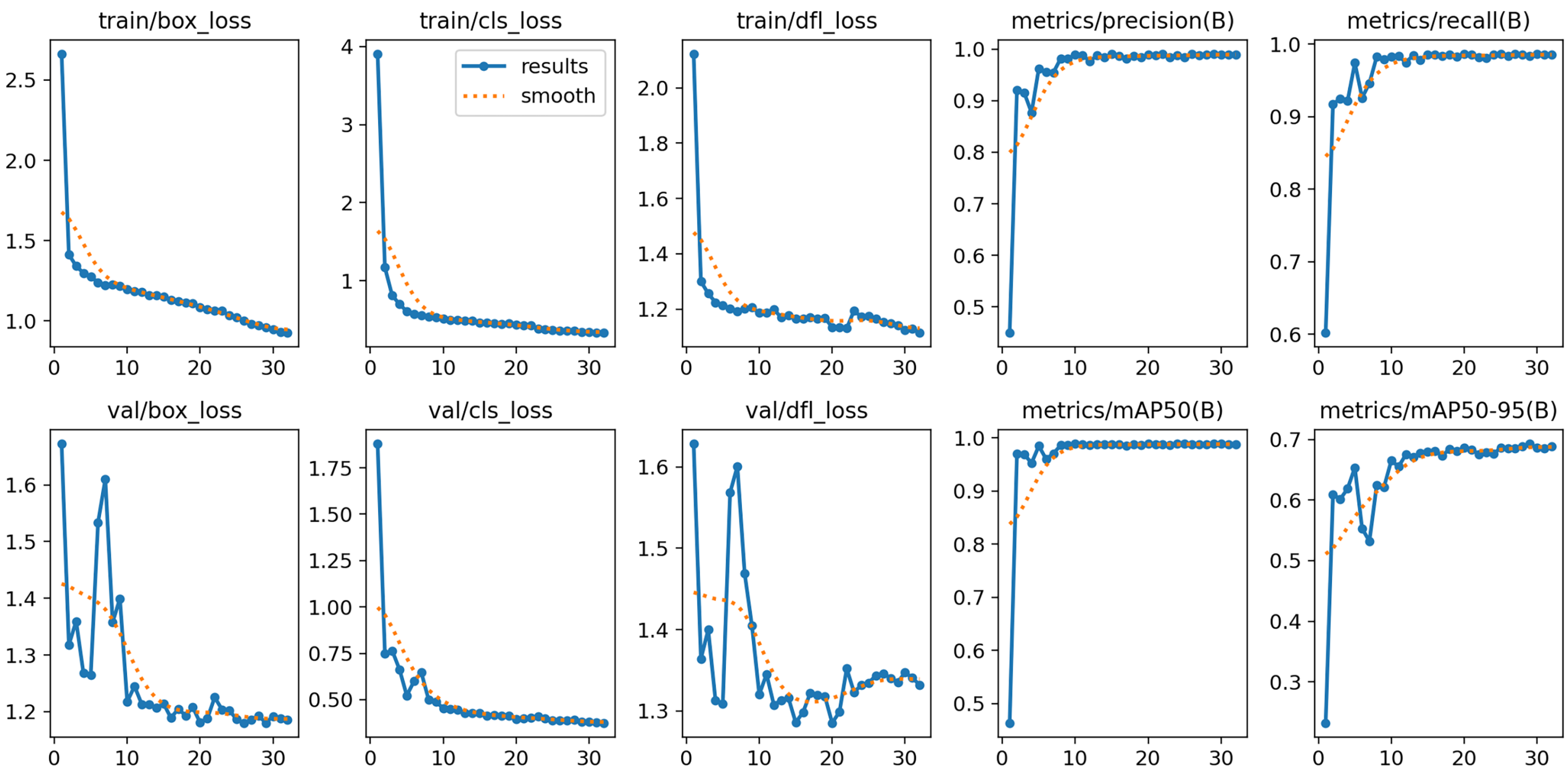

**Figure 12 Training and validation curves of the YOLOv8-small model.** The plots show the convergence of the box, classification, and distribution focal losses along with the trends of precision, recall, and mean average precision (mAP) across 32 epochs.
Full-size DOI: 10.7717/peerj-cs.4103/fig-12

The training and validation loss curves presented in Fig. 12 show a steady decline over epochs, indicating stable convergence without serious signs of overfitting issues. Similarly, the mAP@0.5 and mAP@0.5:0.95 metrics show continuous improvement over the epochs and stabilized around the last epochs. The precision and recall improve rapidly over the first few epochs and remain consistently high thereafter. Therefore, the results are the visual confirmation of the YOLOv8-small model's effective learning and generalization ability. The confusion matrix of Fig. 13 describes how well the model can detect components of unseen circuit diagrams (test dataset). Almost all the classes have been identified successfully with very few mislabelings for visually similar component drawings. This demonstrates that YOLOv8-small generalized well across all classes with minimal cross-class overlap.

## Text region detection

Similar to the component detection, three variants of YOLOv8-nano, small, and medium have been trained for different hyperparameter settings to choose the optimal model. Tables 8 to 10 present a detailed ablation study of three YOLOv8 models with nine key hyperparameters. Just like the case of component detection, higher epoch and image size lead to better accuracy but with an increase in the cost of computation efficiency. Freezing layers will reduce training time as well as model performance. The activation function, RMSprop, has a devastating effect on accuracy, while Stochastic Gradient Descent (SGD)

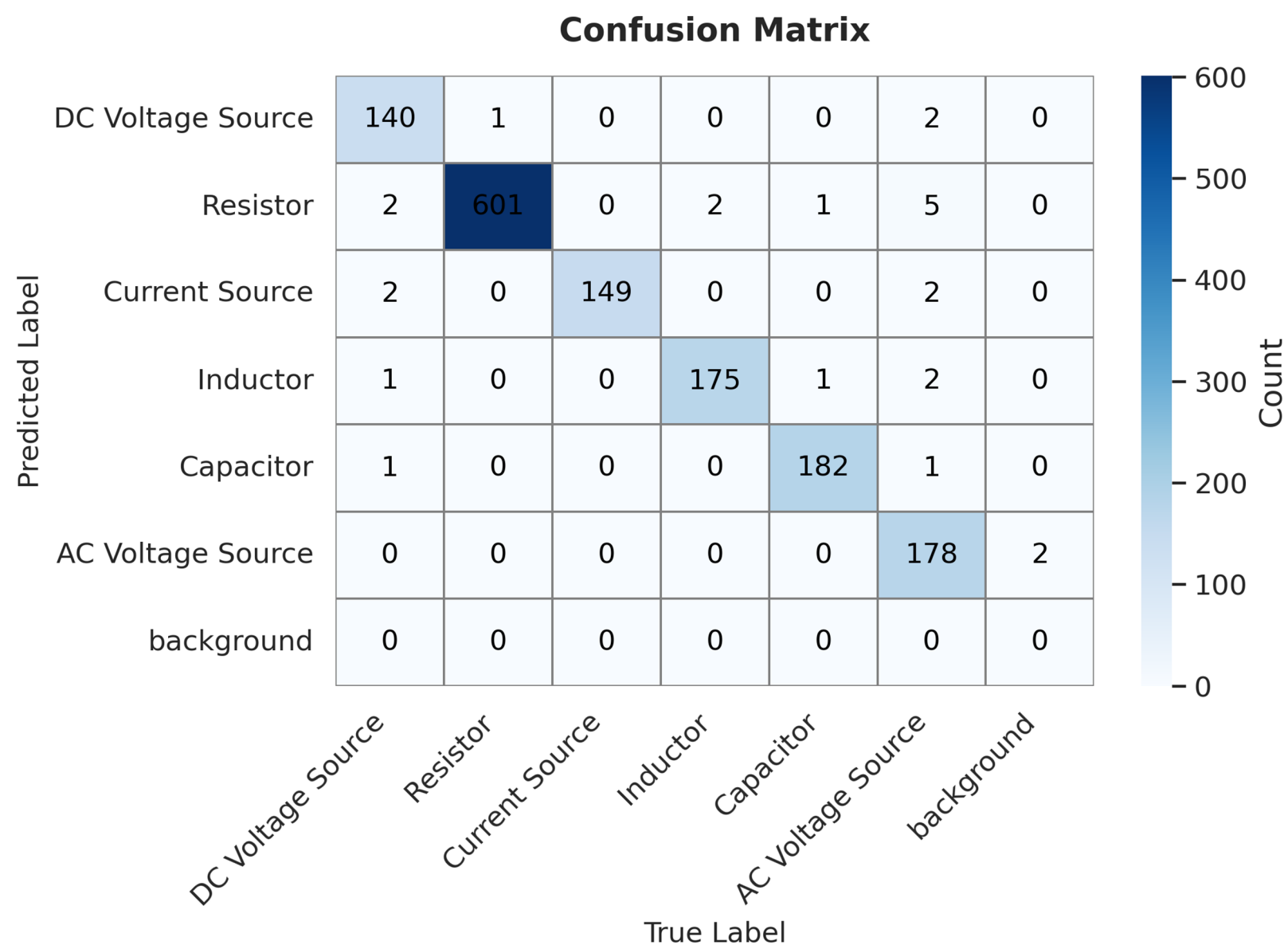

**Figure 13** Confusion matrix of the YOLOv8-small model for component detection, showing strong classwise accuracy with minimal misclassifications across the six component categories.
Full-size DOI: 10.7717/peerj-cs.4103/fig-13

**Table 8** Ablation study of YOLOv8-nano for text region detection.

**Case exploration 1: Modifying epoch**

| Ex. no. | Epoch | Training time | Precision | Recall | mAP@0.5 | mAP@0.5:0.95 | Discovering |
|---|---|---|---|---|---|---|---|
| 1 | 16 | 238.37 | 0.99721 | 0.99478 | 0.99465 | 0.70744 | Lower mAP@0.5:0.95 |
| 2 | 32 | 458.6 | 0.99622 | 0.99444 | 0.99477 | 0.73197 | Slightly lower mAP |
| 3 | 64 | 902.3 | 0.99784 | 0.99333 | 0.99432 | 0.75655 | Best performance |

**Case exploration 2: Modifying weight decay**

| Ex. no. | Weight decay | Training time | Precision | Recall | mAP@0.5 | mAP@0.5:0.95 | Discovering |
|---|---|---|---|---|---|---|---|
| 1 | 0 | 458.41 | 0.99622 | 0.99444 | 0.9947 | 0.73197 | Baseline |
| 2 | 0.0001 | 458.25 | 0.99622 | 0.99444 | 0.9947 | 0.73197 | No change |
| 3 | 0.0005 | 460.87 | 0.99622 | 0.99444 | 0.9947 | 0.73197 | No change |
| 4 | 0.001 | 463.15 | 0.99622 | 0.99444 | 0.9947 | 0.73197 | No change |
| 5 | 0.005 | 463.39 | 0.99622 | 0.99444 | 0.9947 | 0.73197 | No change |

**Case exploration 3: Modifying batch size**

| Ex. no. | Batch size | Training time | Precision | Recall | mAP@0.5 | mAP@0.5:0.95 | Discovering |
|---|---|---|---|---|---|---|---|
| 1 | 16 | 515.74 | 0.99389 | 0.99555 | 0.99444 | 0.73239 | Highest mAP@0.5:0.95 |
| 2 | 32 | 465.39 | 0.99622 | 0.99444 | 0.99473 | 0.73197 | Balanced baseline |
| 3 | 64 | 449.77 | 0.99545 | 0.99605 | 0.99455 | 0.73081 | Improved stability |

**Case exploration 4: Modifying dropout rate**

| Ex. no. | Dropout rate | Training time | Precision | Recall | mAP@0.5 | mAP@0.5:0.95 | Discovering |
|---|---|---|---|---|---|---|---|
| 1 | 0 | 471.56 | 0.99622 | 0.99444 | 0.9947 | 0.73197 | Baseline |
| 2 | 0.1 | 464.4 | 0.99622 | 0.99444 | 0.9947 | 0.73197 | No change |

(Continued)

Table 8 (continued)

**Case exploration 4: Modifying dropout rate**

| Ex. no. | Dropout rate | Training time | Precision | Recall | mAP@0.5 | mAP@0.5:0.95 | Discovering |
|---|---|---|---|---|---|---|---|
| 3 | 0.2 | 463.41 | 0.99622 | 0.99444 | 0.9947 | 0.73197 | No change |
| 4 | 0.3 | 457.36 | 0.99622 | 0.99444 | 0.9947 | 0.73197 | No change |
| 5 | 0.5 | 458.67 | 0.99622 | 0.99444 | 0.9947 | 0.73197 | No change |

**Case exploration 5: Modifying image size**

| Ex. no. | Image size | Training time | Precision | Recall | mAP@0.5 | mAP@0.5:0.95 | Discovering |
|---|---|---|---|---|---|---|---|
| 1 | 320 | 226.53 | 0.99776 | 0.99315 | 0.99448 | 0.71656 | Lower mAP@0.5:0.95 |
| 2 | 512 | 348.45 | 0.99701 | 0.99388 | 0.99464 | 0.72444 | Slightly lower mAP |
| 3 | 640 | 456.79 | 0.99622 | 0.99444 | 0.99471 | 0.73197 | Best performance |

**Case exploration 6: Modifying optimizer**

| Ex. no. | Optimizer | Training time | Precision | Recall | mAP@0.5 | mAP@0.5:0.95 | Discovering |
|---|---|---|---|---|---|---|---|
| 1 | SGD | 457.57 | 0.99622 | 0.99444 | 0.9947 | 0.73197 | Balanced baseline |
| 2 | Adam | 460.64 | 0.99253 | 0.99388 | 0.99452 | 0.71447 | Slightly lower mAP |
| 3 | AdamW | 463.93 | 0.99375 | 0.99499 | 0.99433 | 0.72401 | Slightly lower mAP |
| 4 | RMSProp | 463.36 | 0.96645 | 0.94514 | 0.97844 | 0.54309 | Accuracy dropped |
| 5 | Adamax | 458.61 | 0.99553 | 0.99388 | 0.99399 | 0.73196 | Best overall |

**Case exploration 7: Modifying learning rate**

| Ex. no. | Learning rate | Training time | Precision | Recall | mAP@0.5 | mAP@0.5:0.95 | Discovering |
|---|---|---|---|---|---|---|---|
| 1 | 0.001 | 483.19 | 0.99727 | 0.99333 | 0.99369 | 0.70411 | Lower mAP |
| 2 | 0.005 | 500.17 | 0.99666 | 0.99532 | 0.99466 | 0.72074 | Slightly lower mAP |
| 3 | 0.01 | 494.95 | 0.99622 | 0.99444 | 0.99470 | 0.73197 | Best performance |
| 4 | 0.05 | 493.74 | 0.99392 | 0.99444 | 0.99389 | 0.73307 | Best performance |

**Case exploration 8: Freezing layers**

| Ex. no. | Freezing layers | Training time | Precision | Recall | mAP@0.5 | mAP@0.5:0.95 | Discovering |
|---|---|---|---|---|---|---|---|
| 1 | 0 | 513.66 | 0.99622 | 0.99444 | 0.9947 | 0.73197 | Best performance |
| 2 | 10 | 452.43 | 0.99376 | 0.99333 | 0.99165 | 0.69899 | Accuracy stable |
| 3 | 15 | 455.12 | 0.98482 | 0.99055 | 0.99111 | 0.68906 | Accuracy dropped |
| 4 | 22 | 439.2 | 0.95175 | 0.96107 | 0.98368 | 0.60693 | Accuracy dropped |

**Case exploration 9: Activation function**

| Ex. no. | Activation function | Training time | Precision | Recall | mAP@0.5 | mAP@0.5:0.95 | Discovering |
|---|---|---|---|---|---|---|---|
| 1 | ReLU | 452.9 | 0.98989 | 0.99055 | 0.99332 | 0.69765 | Accuracy dropped |
| 2 | Mish | 462.72 | 0.99476 | 0.98943 | 0.99451 | 0.70015 | Stable performance |
| 3 | LeakyReLU | 454.44 | 0.99188 | 0.99055 | 0.99349 | 0.69943 | Accuracy dropped |
| 4 | SiLU | 457.16 | 0.99372 | 0.99166 | 0.99385 | 0.70581 | Stable performance |

showed stable performance in all the cases. Increased batch size and dropout rate have no significant impact on model accuracy, but it reduces the training time.

Based on all these hyperparameter analyses, the optimal configurations of all the YOLOv8 variants are presented in Table 11. The final nano, small, and medium models are

**Table 9** Ablation study of YOLOv8-small for text region detection.

**Case exploration 1: Modifying epoch**

| Ex. no. | Epoch | Training time | Precision | Recall | mAP@0.5 | mAP@0.5:0.95 | Discovering |
|---|---|---|---|---|---|---|---|
| 1 | 16 | 384.48 | 0.99698 | 0.99555 | 0.99470 | 0.73506 | Lower mAP@0.5:0.95 |
| 2 | 32 | 729.55 | 0.99611 | 0.99607 | 0.99466 | 0.75823 | Slightly lower mAP |
| 3 | 64 | 1,440.36 | 0.99777 | 0.99636 | 0.99468 | 0.77595 | Best performance |

**Case exploration 2: Modifying weight decay**

| Ex. no. | Weight decay | Training time | Precision | Recall | mAP@0.5 | mAP@0.5:0.95 | Discovering |
|---|---|---|---|---|---|---|---|
| 1 | 0 | 746.38 | 0.99611 | 0.99607 | 0.99466 | 0.75823 | Baseline |
| 2 | 0.0001 | 730.5 | 0.99611 | 0.99607 | 0.99466 | 0.75823 | No change |
| 3 | 0.0005 | 730.93 | 0.99611 | 0.99607 | 0.99466 | 0.75823 | No change |
| 4 | 0.001 | 733.49 | 0.99611 | 0.99607 | 0.99466 | 0.75823 | No change |
| 5 | 0.005 | 732.03 | 0.99611 | 0.99607 | 0.99466 | 0.75823 | No change |

**Case exploration 3: Modifying batch size**

| Ex. no. | Batch size | Training time | Precision | Recall | mAP@0.5 | mAP@0.5:0.95 | Discovering |
|---|---|---|---|---|---|---|---|
| 1 | 16 | 809.12 | 0.99538 | 0.99666 | 0.99426 | 0.75856 | Highest mAP@0.5:0.95 |
| 2 | 32 | 734.00 | 0.99611 | 0.99607 | 0.99466 | 0.75823 | Balanced baseline |
| 3 | 64 | 725.33 | 0.99765 | 0.99611 | 0.99454 | 0.75754 | Improved stability |

**Case exploration 4: Modifying dropout rate**

| Ex. no. | Dropout rate | Training time | Precision | Recall | mAP@0.5 | mAP@0.5:0.95 | Discovering |
|---|---|---|---|---|---|---|---|
| 1 | 0 | 741.31 | 0.99611 | 0.99607 | 0.99466 | 0.75823 | Baseline |
| 2 | 0.1 | 731.61 | 0.99611 | 0.99607 | 0.99466 | 0.75823 | No change |
| 3 | 0.2 | 731.05 | 0.99611 | 0.99607 | 0.99466 | 0.75823 | No change |
| 4 | 0.3 | 732.56 | 0.99611 | 0.99607 | 0.99466 | 0.75823 | No change |
| 5 | 0.5 | 733.02 | 0.99611 | 0.99607 | 0.99466 | 0.75823 | No change |

**Case exploration 5: Modifying image size**

| Ex. no. | Image size | Training time | Precision | Recall | mAP@0.5 | mAP@0.5:0.95 | Discovering |
|---|---|---|---|---|---|---|---|
| 1 | 320 | 299.21 | 0.99652 | 0.99499 | 0.99476 | 0.74952 | Lower mAP@0.5:0.95 |
| 2 | 512 | 516.47 | 0.99641 | 0.99555 | 0.99464 | 0.75731 | Slightly lower mAP |
| 3 | 640 | 736.4 | 0.99611 | 0.99607 | 0.99466 | 0.75823 | Best performance |

**Case exploration 6: Modifying optimizer**

| Ex. no. | Optimizer | Training time | Precision | Recall | mAP@0.5 | mAP@0.5:0.95 | Discovering |
|---|---|---|---|---|---|---|---|
| 1 | SGD | 732.86 | 0.99611 | 0.99607 | 0.99466 | 0.75823 | Balanced baseline |
| 2 | Adam | 740.61 | 0.99443 | 0.99278 | 0.99353 | 0.70738 | Slightly lower mAP |
| 3 | AdamW | 742.27 | 0.99473 | 0.99110 | 0.99396 | 0.71991 | Slightly lower mAP |
| 4 | RMSProp | 728.73 | 0.66150 | 0.45322 | 0.56881 | 0.14203 | Accuracy dropped |
| 5 | Adamax | 737.96 | 0.99426 | 0.99333 | 0.99411 | 0.73188 | Slightly lower mAP |

**Case Exploration 7: Modifying learning rate**

| Ex. no. | Learning rate | Training time | Precision | Recall | mAP@0.5 | mAP@0.5:0.95 | Discovering |
|---|---|---|---|---|---|---|---|
| 1 | 0.005 | 730.32 | 0.99777 | 0.99601 | 0.99475 | 0.75209 | Slightly lower mAP |

(Continued)

**Table 9** (continued)

| Case Exploration 7: Modifying learning rate | | | | | | | |
|---|---|---|---|---|---|---|---|
| **Ex. no.** | **Learning rate** | **Training time** | **Precision** | **Recall** | **mAP@0.5** | **mAP@0.5:0.95** | **Discovering** |
| 2 | 0.01 | 730.94 | 0.99611 | 0.99607 | 0.99466 | 0.75823 | Best performance |
| 3 | 0.05 | 733.19 | 0.99471 | 0.99555 | 0.99445 | 0.74387 | Slightly lower mAP |
| **Case exploration 8: Freezing layers** | | | | | | | |
| **Ex. no.** | **Freezing layers** | **Training time** | **Precision** | **Recall** | **mAP@0.5** | **mAP@0.5:0.95** | **Discovering** |
| 1 | 0 | 734.65 | 0.99611 | 0.99607 | 0.99466 | 0.75823 | Best performance |
| 2 | 10 | 467.96 | 0.99610 | 0.99550 | 0.99359 | 0.72568 | Accuracy dropped |
| 3 | 15 | 448.19 | 0.98218 | 0.99277 | 0.99271 | 0.71635 | Accuracy dropped |
| 4 | 22 | 391.20 | 0.95834 | 0.97244 | 0.98917 | 0.62428 | Accuracy dropped |
| **Case exploration 9: Activation function** | | | | | | | |
| **Ex. no.** | **Activation function** | **Training time** | **Precision** | **Recall** | **mAP@0.5** | **mAP@0.5:0.95** | **Discovering** |
| 1 | ReLU | 730.47 | 0.98763 | 0.98554 | 0.99223 | 0.66463 | Accuracy dropped |
| 2 | Mish | 742.25 | 0.99031 | 0.99333 | 0.99411 | 0.71615 | Stable performance |
| 3 | LeakyReLU | 721.52 | 0.99111 | 0.99191 | 0.99402 | 0.69599 | Accuracy dropped |
| 4 | SiLU | 730.49 | 0.99581 | 0.99115 | 0.99414 | 0.70111 | Stable performance |

**Table 10** Ablation study of YOLOv8-medium for text region detection.

| Case exploration 1: Modifying epoch | | | | | | | |
|---|---|---|---|---|---|---|---|
| **Ex. no.** | **Epoch** | **Training time** | **Precision** | **Recall** | **mAP@0.5** | **mAP@0.5:0.95** | **Discovering** |
| 1 | 16 | 763.86 | 0.99663 | 0.99388 | 0.99412 | 0.74142 | Lower mAP@0.5:0.95 |
| 2 | 32 | 1,474.82 | 0.99606 | 0.99778 | 0.99467 | 0.76455 | Slightly lower mAP |
| 3 | 64 | 2,931.67 | 0.99653 | 0.99555 | 0.99452 | 0.78141 | **Best performance** |
| **Case exploration 2: Modifying weight decay** | | | | | | | |
| **Ex. no.** | **Weight decay** | **Training time** | **Precision** | **Recall** | **mAP@0.5** | **mAP@0.5:0.95** | **Discovering** |
| 1 | 0 | 1,472.6 | 0.99606 | 0.99778 | 0.99467 | 0.76455 | Baseline |
| 2 | 0.0001 | 1,470.4 | 0.99606 | 0.99778 | 0.99467 | 0.76455 | No change |
| 3 | 0.0005 | 1,474.8 | 0.99606 | 0.99778 | 0.99467 | 0.76455 | No change |
| 4 | 0.001 | 1,476.2 | 0.99606 | 0.99778 | 0.99467 | 0.76455 | No change |
| 5 | 0.005 | 1,475.3 | 0.99606 | 0.99778 | 0.99467 | 0.76455 | No change |
| **Case exploration 3: Modifying batch size** | | | | | | | |
| **Ex. no.** | **Batch size** | **Training time** | **Precision** | **Recall** | **mAP@0.5** | **mAP@0.5:0.95** | **Discovering** |
| 1 | 16 | 1,637.85 | 0.99598 | 0.99666 | 0.99480 | 0.76450 | Highest mAP@0.5:0.95 |
| 2 | 32 | 1,477.32 | 0.99606 | 0.99778 | 0.99467 | 0.76455 | Balanced baseline |
| 3 | 64 | 1,380.12 | 0.99701 | 0.99612 | 0.99451 | 0.76288 | Improved stability |
| **Case exploration 4: Modifying dropout rate** | | | | | | | |
| **Ex. no.** | **Dropout rate** | **Training time** | **Precision** | **Recall** | **mAP@0.5** | **mAP@0.5:0.95** | **Discovering** |
| 1 | 0 | 1,476.3 | 0.99606 | 0.99778 | 0.99467 | 0.76455 | Baseline |
| 2 | 0.1 | 1,469.5 | 0.99606 | 0.99778 | 0.99467 | 0.76455 | No change |

**Table 10** (continued)

**Case exploration 4: Modifying dropout rate**

| Ex. no. | Dropout rate | Training time | Precision | Recall | mAP@0.5 | mAP@0.5:0.95 | Discovering |
|---|---|---|---|---|---|---|---|
| 3 | 0.2 | 1,468.8 | 0.99606 | 0.99778 | 0.99467 | 0.76455 | No change |
| 4 | 0.3 | 1,471.7 | 0.99606 | 0.99778 | 0.99467 | 0.76455 | No change |
| 5 | 0.5 | 1,472.1 | 0.99606 | 0.99778 | 0.99467 | 0.76455 | No change |

**Case exploration 5: Modifying image size**

| Ex. no. | Image size | Training time | Precision | Recall | mAP@0.5 | mAP@0.5:0.95 | Discovering |
|---|---|---|---|---|---|---|---|
| 1 | 320 | 728.45 | 0.99641 | 0.99555 | 0.99464 | 0.74910 | Lower mAP@0.5:0.95 |
| 2 | 512 | 1,105.22 | 0.99622 | 0.99600 | 0.99466 | 0.75890 | Slightly lower mAP |
| 3 | 640 | 1,474.82 | 0.99606 | 0.99778 | 0.99467 | 0.76455 | Best performance |

**Case exploration 6: Modifying optimizer**

| Ex. no. | Optimizer | Training time | Precision | Recall | mAP@0.5 | mAP@0.5:0.95 | Discovering |
|---|---|---|---|---|---|---|---|
| 1 | SGD | 1,474.8 | 0.99606 | 0.99778 | 0.99467 | 0.76455 | Balanced baseline |
| 2 | Adam | 1,491.3 | 0.99412 | 0.99487 | 0.99393 | 0.73121 | Slightly lower mAP |
| 3 | AdamW | 1,498.2 | 0.99501 | 0.99512 | 0.99421 | 0.74209 | Stable performance |
| 4 | RMSProp | 1,460.3 | 0.68555 | 0.47124 | 0.57391 | 0.15478 | Accuracy dropped |
| 5 | Adamax | 1,477.1 | 0.99587 | 0.99499 | 0.99441 | 0.75542 | Slightly lower mAP |

**Case exploration 7: Modifying learning rate**

| Ex. no. | Learning rate | Training time | Precision | Recall | mAP@0.5 | mAP@0.5:0.95 | Discovering |
|---|---|---|---|---|---|---|---|
| 1 | 0.005 | 1,470.4 | 0.99711 | 0.99601 | 0.99475 | 0.75250 | Slightly lower mAP |
| 2 | 0.01 | 1,474.8 | 0.99606 | 0.99778 | 0.99467 | 0.76455 | Best performance |
| 3 | 0.05 | 1,479.9 | 0.99492 | 0.99512 | 0.99445 | 0.74321 | Slightly lower mAP |

**Case exploration 8: Freezing layers**

| Ex. no. | Freezing layers | Training time | Precision | Recall | mAP@0.5 | mAP@0.5:0.95 | Discovering |
|---|---|---|---|---|---|---|---|
| 1 | 0 | 1,477.3 | 0.99606 | 0.99778 | 0.99467 | 0.76455 | Best performance |
| 2 | 10 | 992.6 | 0.99510 | 0.99522 | 0.99391 | 0.73598 | Accuracy stable |
| 3 | 15 | 951.2 | 0.98315 | 0.99277 | 0.99271 | 0.72435 | Accuracy dropped |
| 4 | 22 | 888.7 | 0.95634 | 0.97244 | 0.98917 | 0.62128 | Accuracy dropped |

**Case exploration 9: Activation function**

| Ex. no. | Activation function | Training time | Precision | Recall | mAP@0.5 | mAP@0.5:0.95 | Discovering |
|---|---|---|---|---|---|---|---|
| 1 | ReLU | 1,469.9 | 0.98893 | 0.98554 | 0.99233 | 0.66751 | Accuracy dropped |
| 2 | Mish | 1,491.6 | 0.99191 | 0.99333 | 0.99422 | 0.72011 | Stable performance |
| 3 | LeakyReLU | 1,458.1 | 0.99255 | 0.99191 | 0.99401 | 0.70398 | Accuracy dropped |
| 4 | SiLU | 1,474.8 | 0.99606 | 0.99778 | 0.99467 | 0.76455 | Best |

**Table 11** The best hyperparameter settings for the final nano, small and medium models.

| Model name | Epoch | Weight decay | Batch size | Dropout | Image size | Optimizer | Learning rate | Freezing layers | Activation function |
|---|---|---|---|---|---|---|---|---|---|
| Nano | 64 | 0.001 | 16 | 0.0 | 640 | SGD | 0.05 | 0 | SiLU |
| Small | 64 | 0.0005 | 32 | 0.0 | 640 | SGD | 0.01 | 0 | SiLU |
| Medium | 32 | 0.001 | 32 | 0.0 | 640 | SGD | 0.01 | 0 | SiLU |

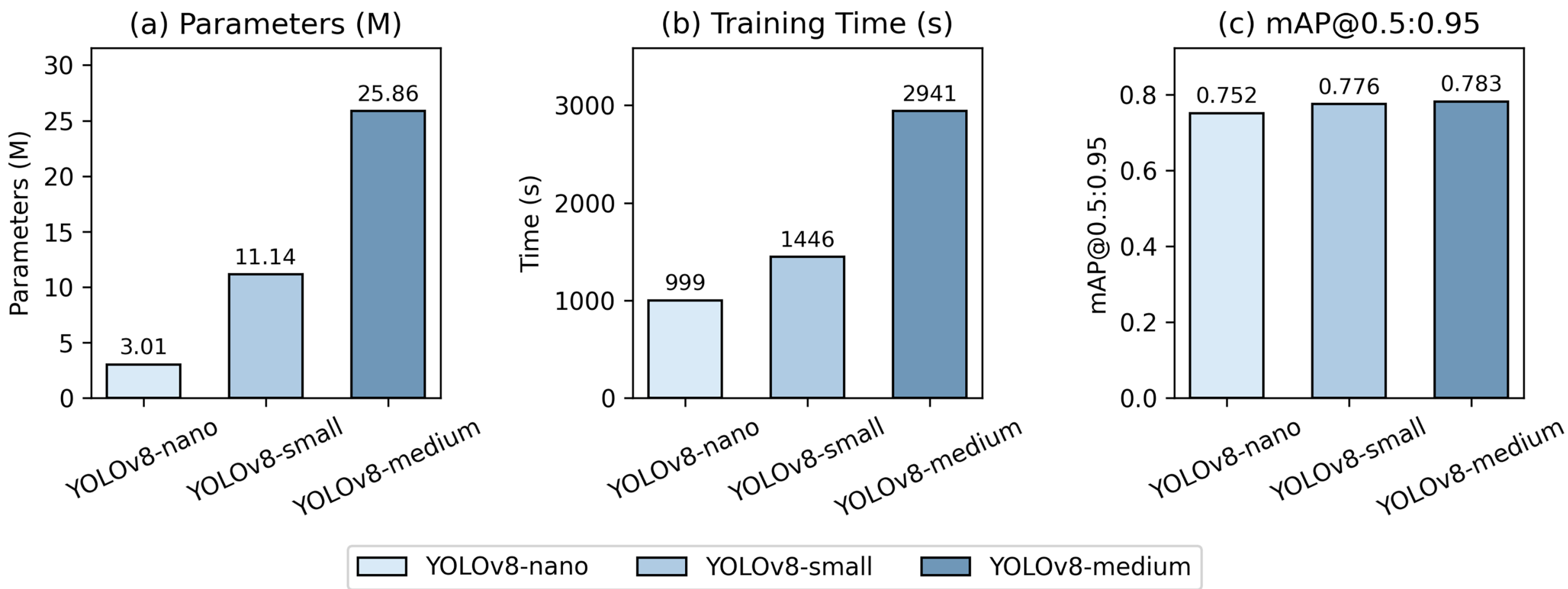

**Figure 14** (A–C) YOLOv8 nano, small, and medium models' best parameter evaluation for the text-region localization task.
Full-size DOI: 10.7717/peerj-cs.4103/fig-14

trained according to these settings. The performance comparison of the final YOLOv8 models is presented in Fig. 14, which affirms the superiority of the YOLOv8-small variant over the nano and medium variants. The 3.2% mAP improvements with a cost of almost 30% increased training time and 3.7× more parameters of the 'small' variant when compared to 'nano' is critical. A failure in text-region detection is an unrecoverable error for the text recognition. Therefore, without relying on a faster variant, a more stable variant is chosen. Although YOLOv8-medium has the highest observed accuracy, the numeric difference is marginal when compared to the small variant. However, YOLOv8-medium is 56% bigger and requires almost 2× more time than YOLOv8-small. Similar to component detection, the performance differences between YOLO variants are reported in Fig. 14 descriptively due to the limited dataset size. Therefore, the YOLOv8-small model is selected for text-localization purposes. Similar to the earlier sub-section, the loss and metrics curves of this model are presented in Fig. 15. To account for sampling uncertainty, a 95% confidence interval for the end-to-end accuracy was estimated using a binomial proportion model. Due to the limited evaluation dataset, performance differences between YOLO variants are reported descriptively rather than as statistically significant comparisons.

## Text detection by the CRNN

The text region localized by the YOLOv8-small model was passed to the CRNN model for text recognition. The CRNN model was trained for 50 epochs (Adam, 1e−3). It consists of 1.08 million parameters and achieved a final character error rate (CER) of 6.47%. The training and validation loss curve displayed in Fig. 16 presents a gradual descent where the model was stabilized around 30 epochs, indicating an effective convergence without overfitting. The periodic spikes in validation loss are likely attributable to the inherent high variance of hand-written text. Some examples of the model producing texts from the cropped image are also shown in Fig. 16, from which it is evident that the model can correctly recognize characters and values, yet errors may occur randomly when visually similar characters appear, and these are expected for real-world handwritten datasets.

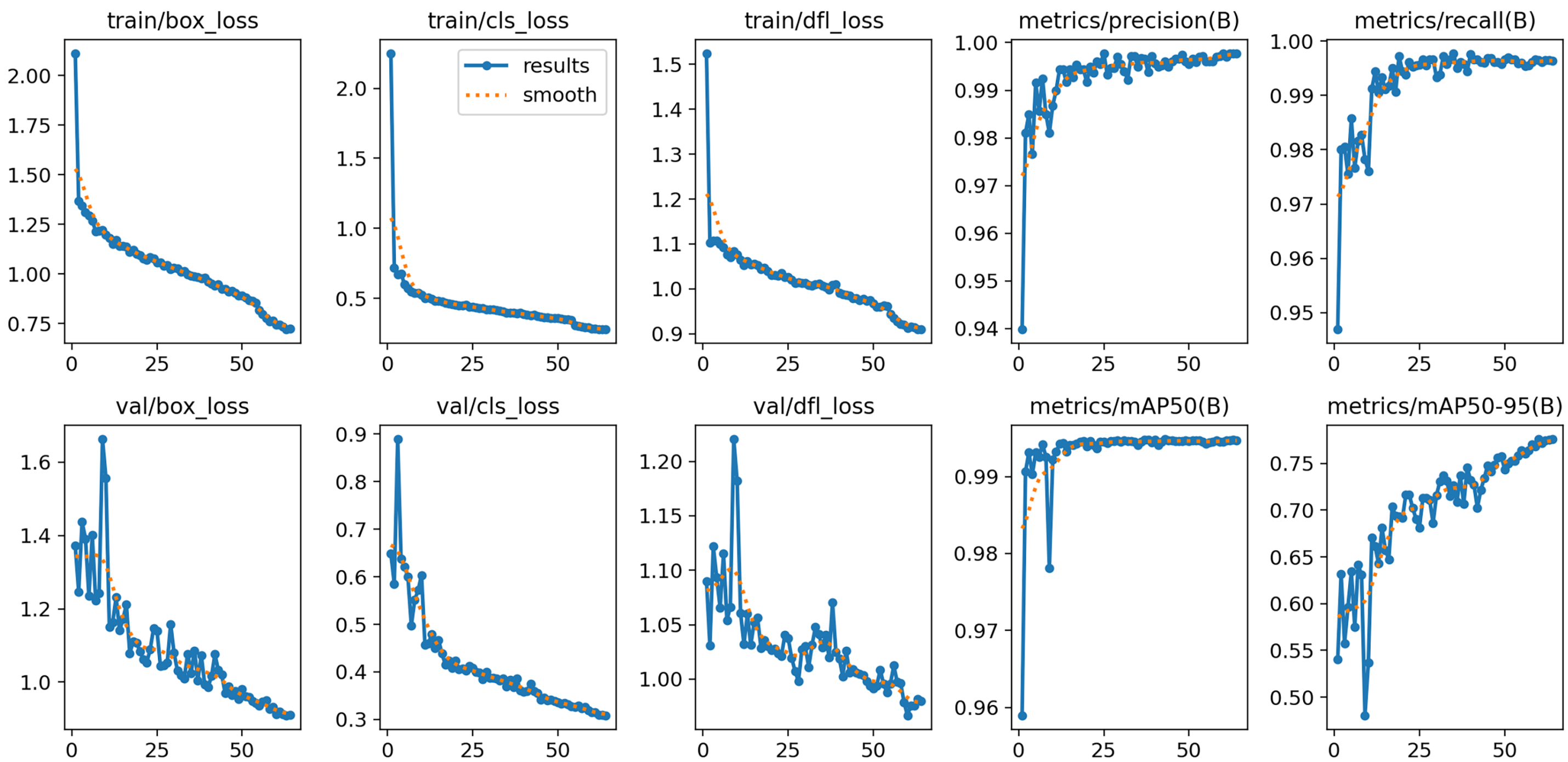

**Figure 15 Training and validation curves of the YOLOv8-small model for text region detection.** The plots illustrate the convergence of the box, classification, and distribution focal losses, alongside the evolution of precision, recall, and mean average precision (mAP) over 64 epochs.
Full-size DOI: 10.7717/peerj-cs.4103/fig-15

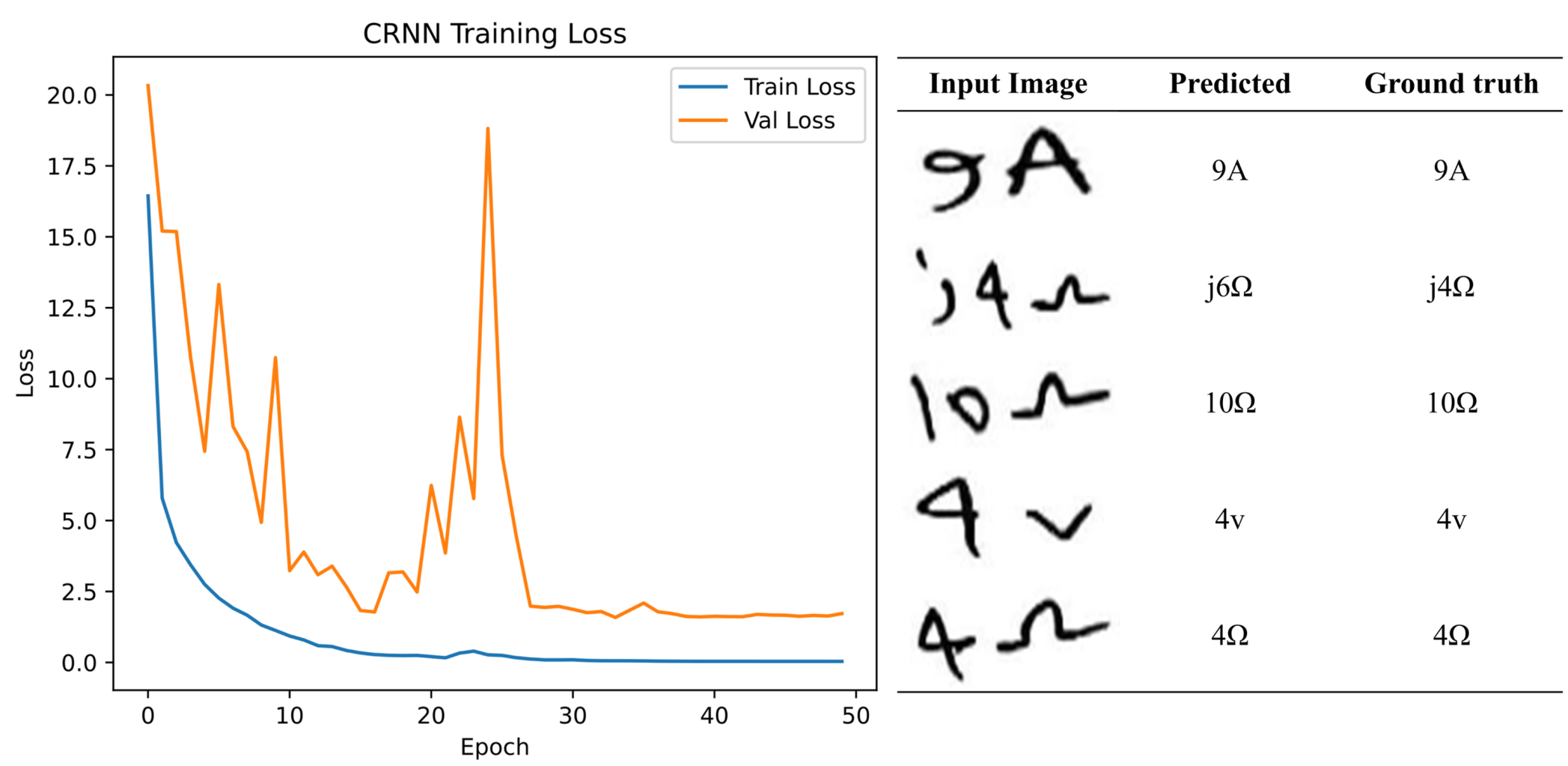

**Figure 16 Training and validation CTC loss curves along with some examples of text extraction of the CRNN model.** Full-size DOI: 10.7717/peerj-cs.4103/fig-16

**Table 12** **Examples of the proposed end-to-end handwritten circuit understanding pipeline.**

| | | | | |
|---|---|---|---|---|
| Prepressed image | 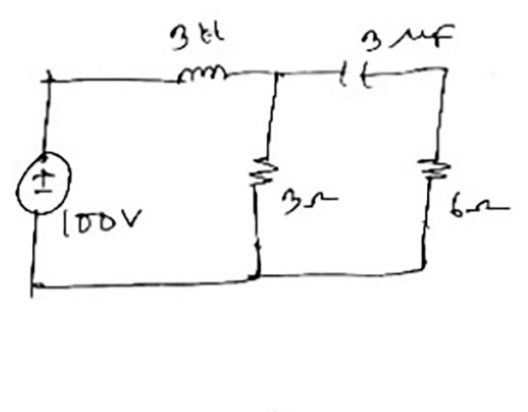 | 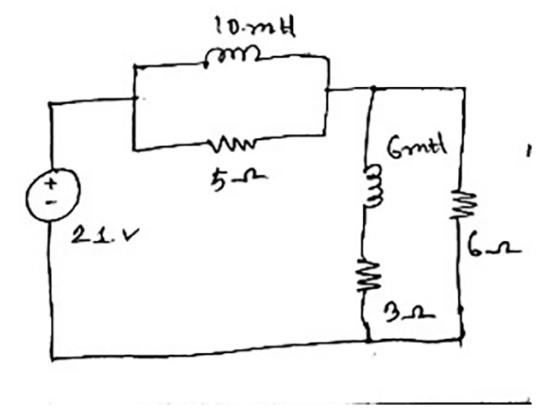 | 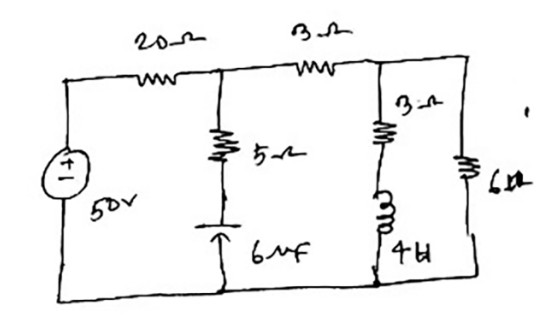 | 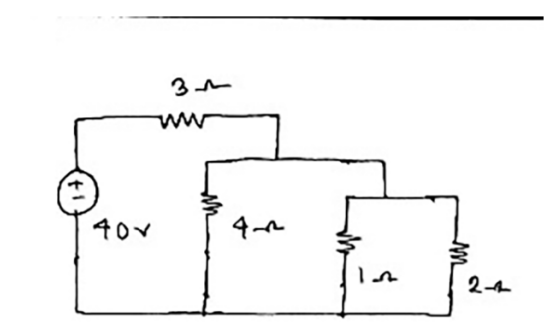 |
| Component detection | 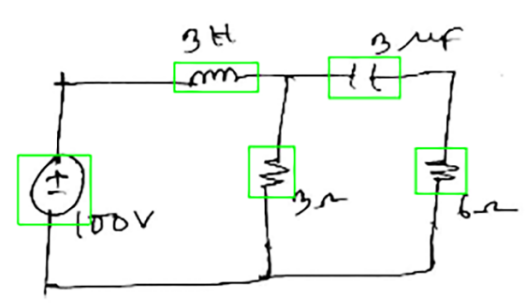 | 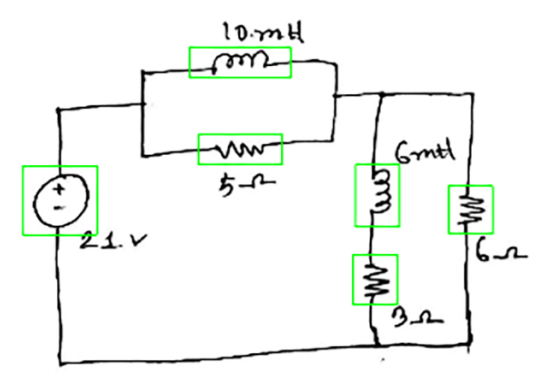 | 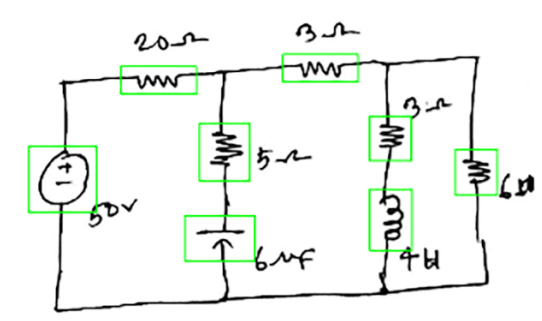 | 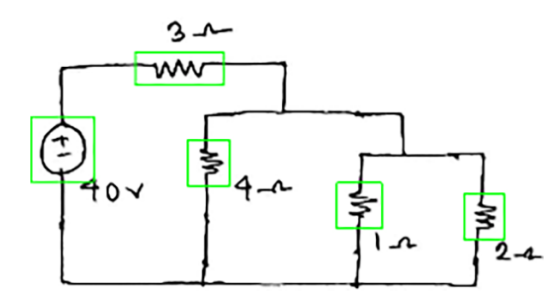 |
| Text detection | 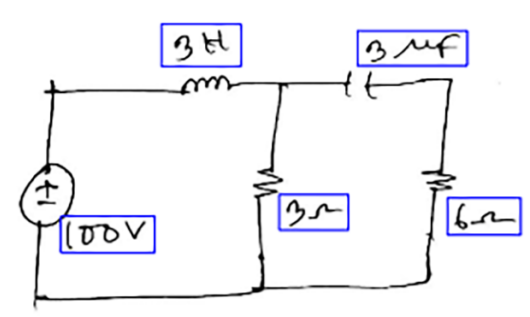 | 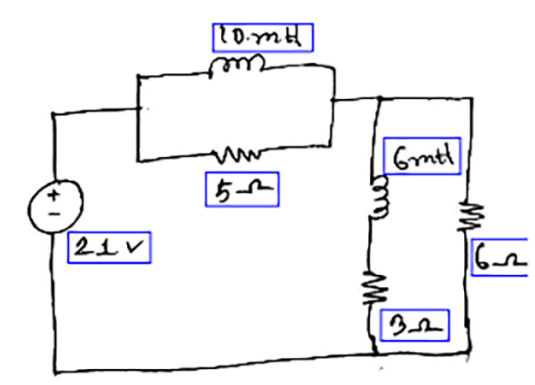 | 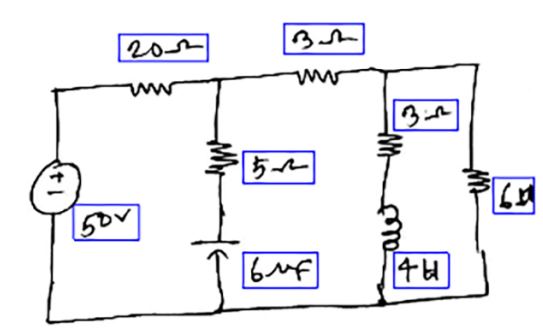 | 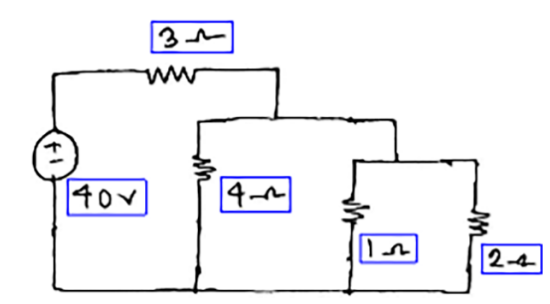 |
| Masked image For node | 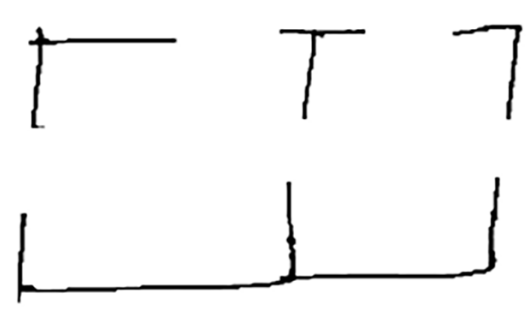 | 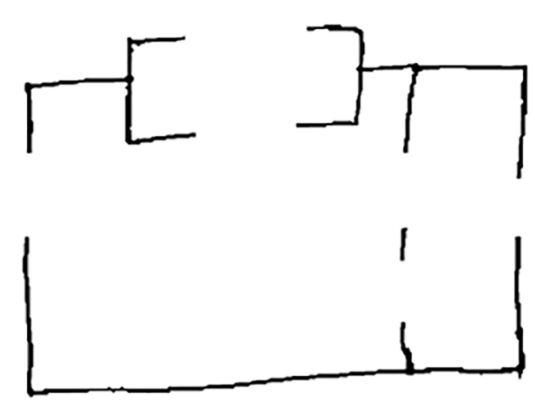 | 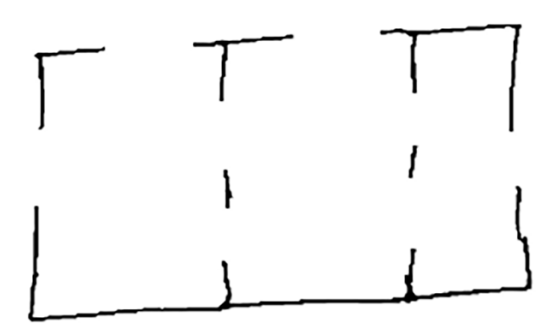 | 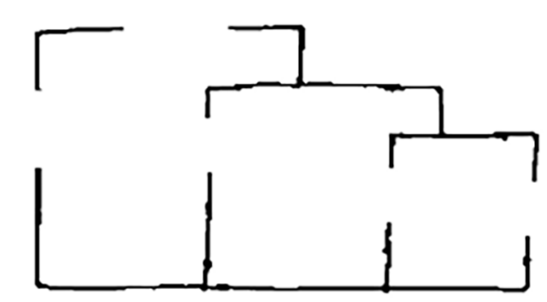 |
| Node detection | 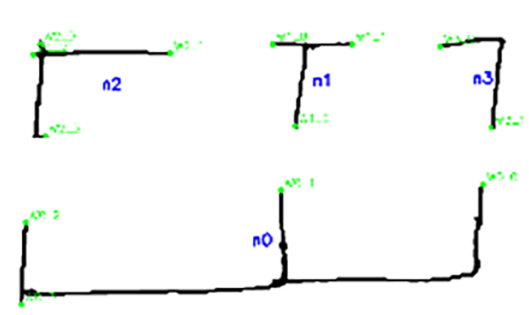 | 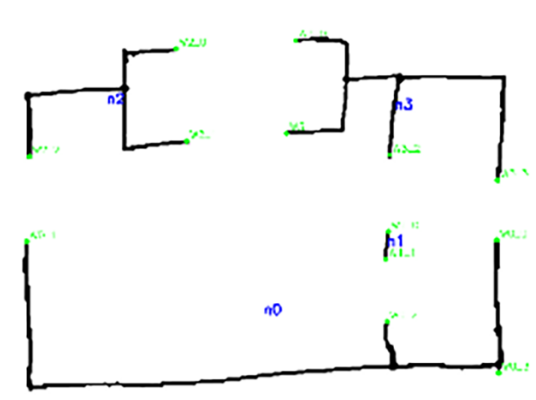 | 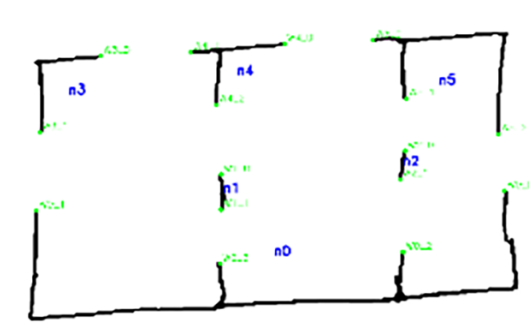 | 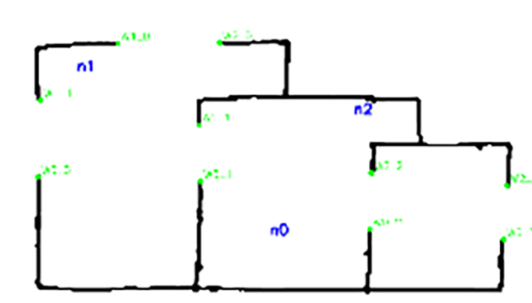 |
| Final annotation | 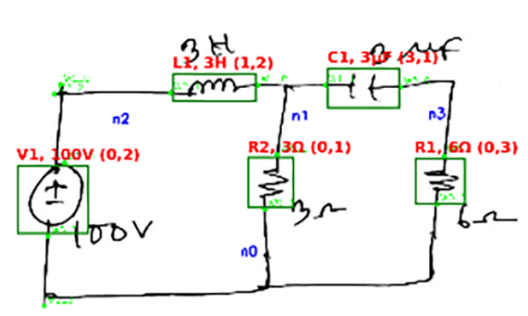 | 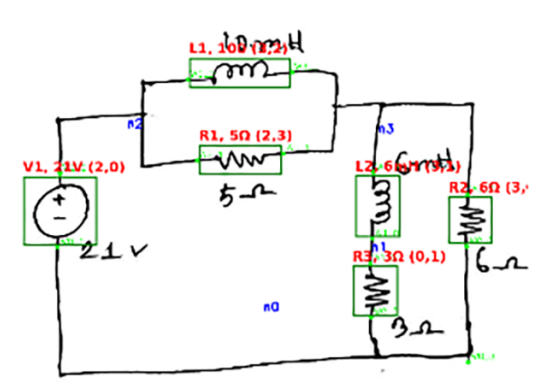 | 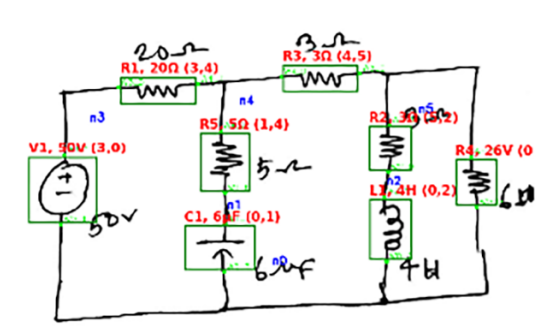 | 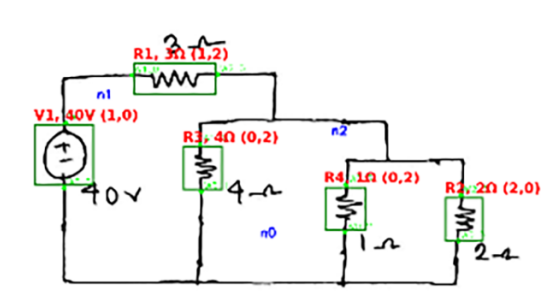 |
| Netlist | C1 3 1 3e-06<br>R1 0 3 6.0<br>R2 0 1 3.0<br>L1 1 2 3.0<br>V1 0 2 DC 100.0 | V1 2 0 DC 21.0<br>R1 2 3 5.0<br>R2 3 0 5.0<br>L1 3 2 10.0<br>L2 3 1 0.002<br>R3 0 1 3.0 | L1 0 2 4.0<br>R1 3 4 20.0<br>C1 0 1 6e-06<br>R2 2 5 3.0<br>R3 4 5 3.0<br>R4 0 5 6.0<br>R5 1 4 5.0<br>V1 0 3 DC 50.0 | V1 1 0 DC 40.0<br>R1 2 1 3.0<br>R2 2 0 2.0<br>R3 0 2 1.0<br>R4 0 2 4.0 |
| Solution | V(3) = 0 V<br>V(1) = 100 V<br>V(2) = 100 V<br>I(V1) = 33.33334 A | V(2) = 21 V<br>V(3) = 21 V<br>V(1) = 21 V<br>I(V1) = 11.2 A | V(2) = 0 V<br>V(3) = 50 V<br>V(4) = 10 V<br>V(1) = 10 V<br>V(5) = 4 V<br>I(V1) = 2 A | V(1) = 40 V<br>V(2) = 6.4 V<br>I(V1) = 11.2 A |

**Table 13** Stress-test behavior on challenging edge cases.

| Edge-case category | No. of images | Component detection fully correct | Component detection partially correct | OCR correct | Topology correct | End-to-end correct | Most common failure |
|---|---|---|---|---|---|---|---|
| Extreme noise, pencil shading | 2 | 2 | – | 1 | 2 | 1 | False contours |
| Overlapping text and symbols | 2 | 1 | 1 | 0 | 1 | 0 | OCR confusion |
| Faint or broken strokes | 2 | 1 | 0 | 1 | 1 | 0 | Missing wire segment |
| Crowded components | 2 | 1 | 1 | 1 | 1 | 0 | Symbol overlap |
| Messy wire intersections | 2 | 0 | 1 | 1 | 0 | 0 | Component missing |
| Complex multi-loop circuits | 2 | 1 | 1 | 1 | 1 | 0 | OCR confusion |

## Node detection, final annotation, netlist generation, and circuit solution

The efficacy of this research relies on solving an electric successfully by integrating all the modules in a single pipeline. Table 12 represents the step-by-step output of the proposed pipeline using four different hand-drawn circuits. Each image has been processed in parallel by the component and text-region detector. Text regions were passed to CRNN to extract component values. A masked binary image was prepared for node detection. All intermediate results are then synthesized to form the final annotated image. Later, a SPICE-style netlist was generated, and nodal analysis was performed to solve the circuits.

## Error and edge-case analysis

### *Edge-case analysis*

To further test the robustness of the proposed pipeline, we test the modular framework with some edge cases such as extreme noise, overlapping text, crowded components, and complex multi-loop RLC circuits. We constructed a stress-test set containing 12 visually and structurally challenging hand-drawn circuits. These samples were intentionally selected to represent difficult cases beyond the distribution of the main evaluation set. Table 13 shows that performance degraded under these conditions, and many examples contained at least one stage-level error. Rather than demonstrating uniform robustness, this analysis is intended to reveal the failure boundaries of the modular pipeline and identify the stages most sensitive to extreme input complexity.

Figure 17 shows representative examples from the stress-test set. As seen from these cases, the framework remains reasonably stable under isolated visual difficulty, but performance degrades when structural complexity and handwritten ambiguity occur together. In particular, complex multi-loop circuits and messy wire intersections mainly affect node mapping and topology reconstruction, whereas overlapping values and overlapping components more often lead to OCR or localization errors.

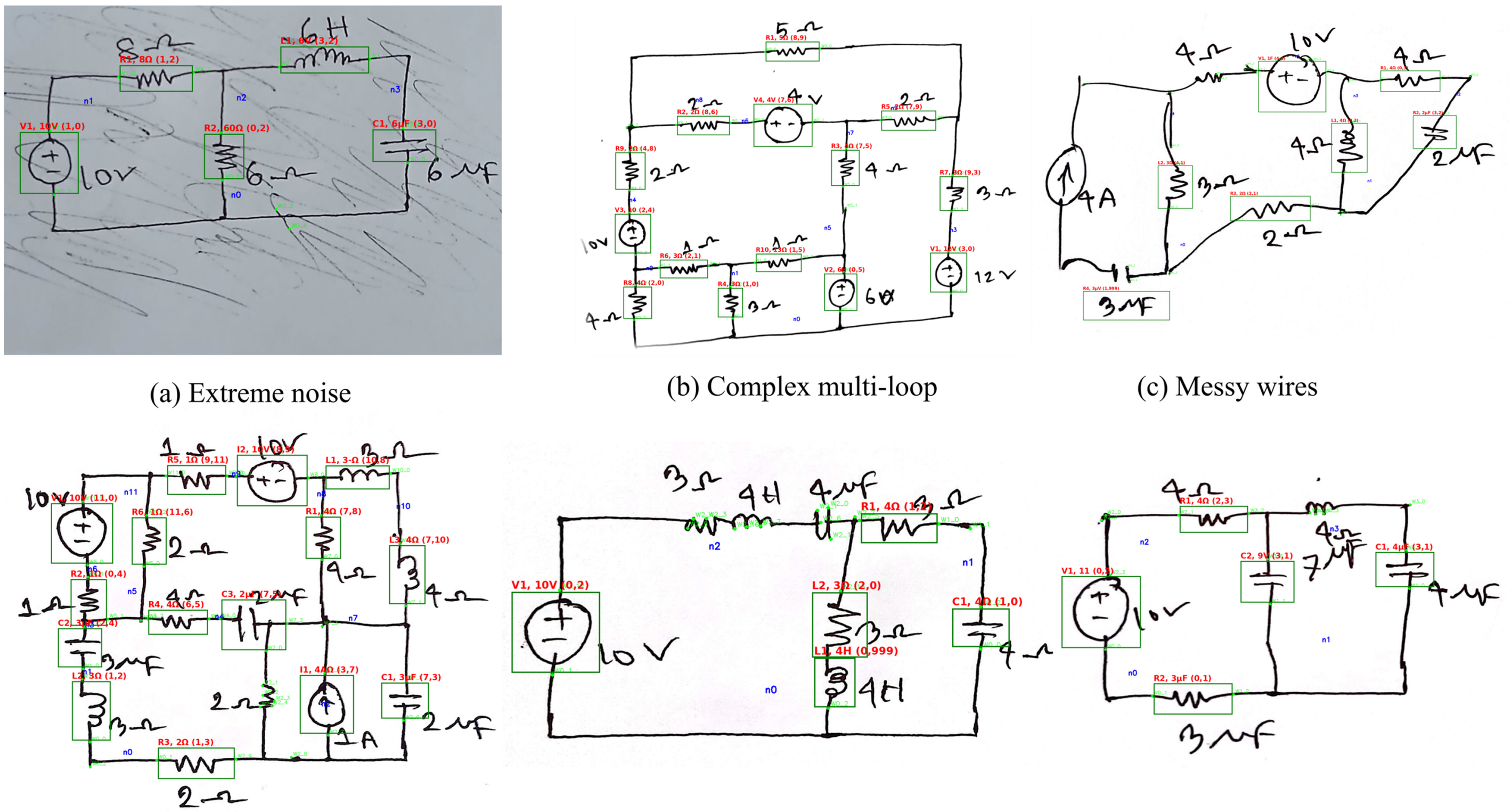

(a) Extreme noise (b) Complex multi-loop (c) Messy wires

(d) Crowded components (e) Overlapping components (f) Overlapping values

**Figure 17** **Representative edge-case stress-test examples: (A) extreme noise, (B) complex multi-loop circuit, (C) messy wires, (D) crowded components, (E) overlapping components, and (F) overlapping values.** Full-size DOI: 10.7717/peerj-cs.4103/fig-17

**Table 14** **Categorized error breakdown on the 50-image ground-truth evaluation set.**

| Error category | Images affected (*n* = 50) | Percentage of all test circuits | Percentage of failed circuits |
|---|---|---|---|
| Component detection | 2 | 4% | 50% |
| Text-region detection | 2 | 4% | 50% |
| OCR/value extraction | 2 | 4% | 50% |
| Topology reconstruction | 3 | 6% | 75% |

### *Categorized end-to-end error breakdown*

To obtain a more interpretable view of failure, we manually audited a standard 50-image evaluation set along with ground-truth and grouped the observed mismatches according to the major stages of the pipeline: component detection, text-region detection, OCR/value extraction, and topology reconstruction. Because a single failed circuit may contain more than one error source, these categories are not mutually exclusive.

Table 14 presents the categorized error breakdown. The results show that topology reconstruction is the most frequent source of failure, affecting three out of 50 test circuits (6%) and 75% of the failed cases. By comparison, component detection, text-region detection, and OCR/value extraction each affected two out of 50 images (4%).
This indicates that the main challenge of the end-to-end framework is not only symbol localization, but also the correct recovery of graph-level connectivity from noisy hand-drawn sketches.

**Table 15** Representative failed cases grouped by dominant error type.

| Error type | Preprocessed image | Error-stage | Final detection | Explanation |
|---|---|---|---|---|
| Component detection | 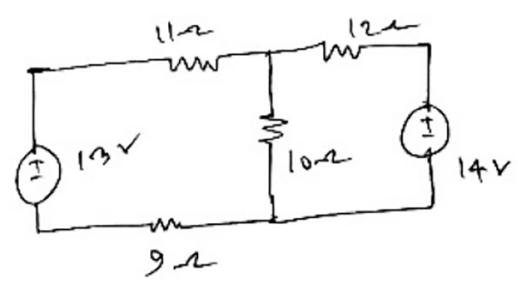 | 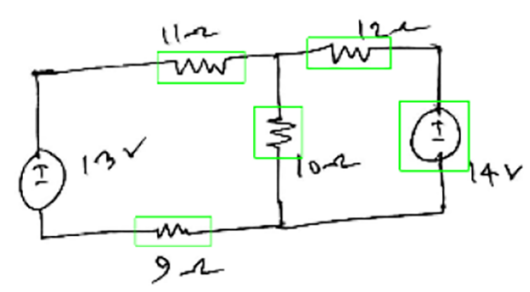 | 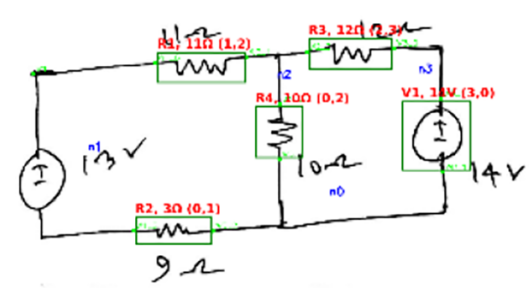 | Missed one voltage source. |
| Text-region detection | 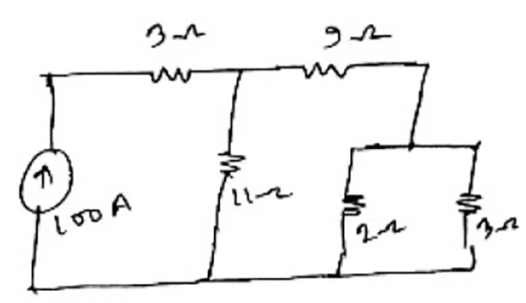 | 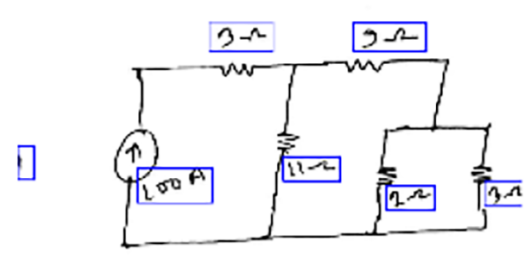 | 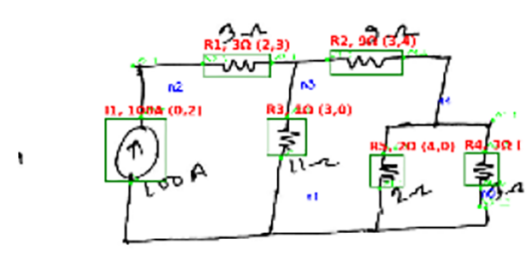 | Falsely identify one noise as text-region. |
| OCR/value extraction | 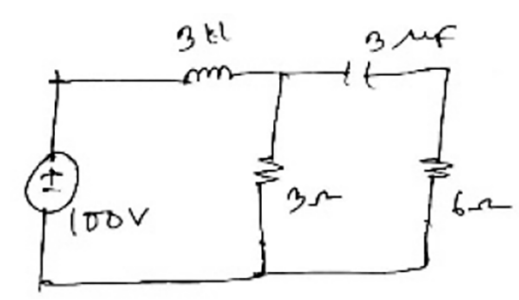 |  1 \| Resistor \| (719 , 275 ) \| 6Ω<br>2 \| Capacitor \| (596 , 129 ) \| 3μF<br>3 \| VoltageSrc \| (112 , 303 ) \| 1000V<br>4 \| Resistor \| (453 , 279 ) \| 3Ω<br>5 \| Inductor \| (364 , 128 ) \| 3H | 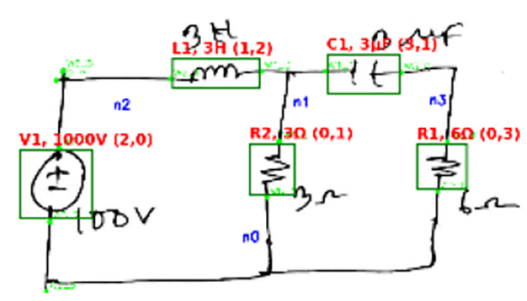 | Falsely extract "100 v" as "1,000 v". |
| Topology reconstruction | 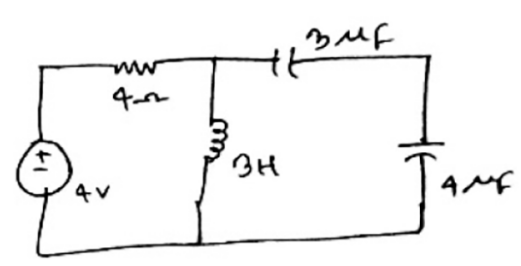 | 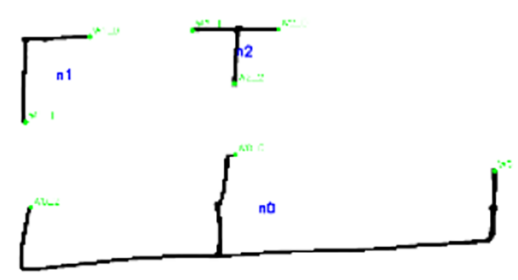 | 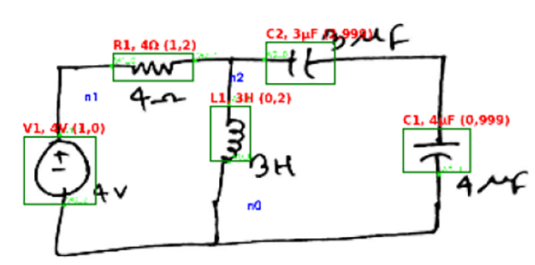 | Missed wire between two capacitor. |

The error distribution indicates that topology-related mistakes are the dominant source of exact-match failure. OCR and text-region errors also contribute meaningfully, particularly when handwritten values are close to component boundaries or are partially merged with surrounding strokes. Component detection errors are less frequent, but when they occur, they usually propagate downstream and affect both value assignment and circuit structure. Table 15 presents one representative example for each major error category: component detection, text-region detection, OCR/value extraction, and topology reconstruction. For each case, the preprocessed input, a zoomed error-stage view, and the final annotated output are shown to visually connect the local cause of failure with its end-to-end effect. Out of the 50 evaluated circuits, the proposed pipeline successfully generated correct end-to-end interpretations for 47 circuits, while three circuits contained at least one stage-level error, resulting in an overall end-to-end 94% accuracy (95% CI [83.5–98.1%]) on the evaluation set.

## Computational complexity and runtime analysis

To evaluate computational efficiency, the runtime of each stage of the proposed pipeline was measured on the standard test dataset consisting of 50 circuit images. Timing was recorded for the major processing stages, including component detection, text detection, topology extraction, OCR-based value assignment, and netlist generation. Visualization and file I/O operations were excluded from the reported timings to measure the algorithmic computational cost. The runtime breakdown of the proposed framework is presented in Table 16. The dominant computational stages are the wire and node extraction module and the OCR-based component value assignment stage. These operations account for the majority of the processing time, whereas netlist generation and circuit solving contribute negligibly to the overall latency. Since each circuit image is processed independently, the proposed framework can also benefit from parallel processing or Graphics Processing Unit (GPU) acceleration in large-scale deployments.

**Table 16** Average computational runtime of each stage of the proposed framework.

| Stage | Mean runtime (ms) |
|---|---|
| Component detection (YOLOv8) | 232.9 |
| Text-region detection (YOLOv8) | 247.9 |
| Wire and node extraction | 706.3 |
| OCR and text-to-component assignment | 750.2 |
| Netlist generation | 0.1 |
| DC circuit analysis | 0.227 |
| **Total compute runtime** | **1,958.4** |

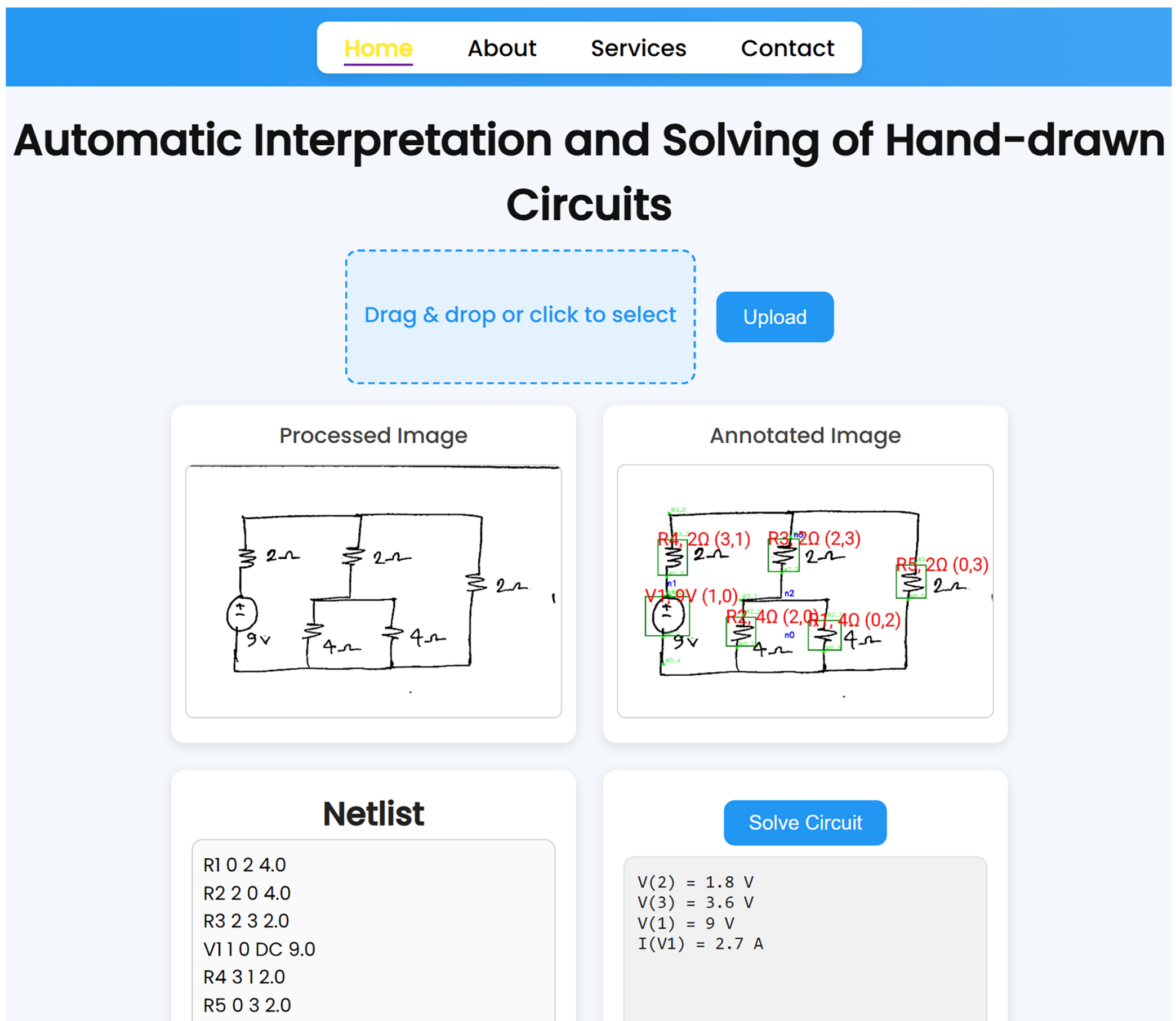

**Figure 18** Illustration of the final working interface of the web application, showing the uploaded circuit, annotated output, netlist, and simulation results displayed together for complete visual and analytical understanding. Full-size DOI: 10.7717/peerj-cs.4103/fig-18

**Table 17** **Comparison with prior studies.**

| Article | Model/Method | Component classes detected | Text/Value detection | Dataset type | Accuracy/mAP |
|---|---|---|---|---|---|
| *Alhalabi et al. (2021)* | Capsule Network (CapsNet) + Smartphone App | **Five classes:** Resistor, Inductor, Capacitor, DC Voltage Source, Current Source | Manual input for component values | Hand-drawn circuits captured *via* smartphone | mAP ≈ 80.61%, IoU 0.5 |
| *Rachala & Panicker (2022)* | YOLOv5 + Node Recognition (Graph Mapping) | **Six classes:** Resistor, Capacitor, Inductor, DC Source, AC Source, Current Source | Absent | Hand-drawn + printed hybrid dataset | mAP@0.5 = 98.2%, overall = 80% |
| **This work** | YOLOv8 + CRNN + DBSCAN (End-to-End Pipeline) + Web Implementation | **Six classes:** Resistor, Capacitor, Inductor, DC Voltage Source, AC Voltage Source, Current Source | CRNN-CTC OCR for handwritten values (*e.g.*, 10 kΩ, 5 μF) | 1,000 hand-drawn diagrams (custom dataset) | mAP@0.5 = 98.9%, mAP@0.5:0.95 = 69.7%, CER = 6.47%, overall = 94% |

## Web implementation

To make the proposed framework accessible and interactive, a web-based interface has been developed, which is depicted in Fig. 18. The backend was built on Flask (Python Software Foundation, Beaverton, OR, USA), and the frontend was created using HTML, CSS, and JavaScript. The design emphasized simplicity and user-friendliness. Once the image is uploaded, there is a drag-and-drop option. The system receives the image and executes the entire framework at the backend, and returns two visual results: a netlist and a final annotated image with detected components, values, and node connections highlighted. The user can edit or verify the netlist if necessary. Finally, by clicking the solve button, the system performs nodal analysis and presents calculated voltages and currents instantly. It's lightweight and works on any browser.

## Comparison with prior studies

To the best of our knowledge, context-wise, only a limited number of studies have addressed the problem of automating hand-written RLC analog circuit simulation, moving a step ahead of just detecting components. A detailed comparison with those state-of-the-art studies is discussed here to validate the proposed framework's contribution. Although *Alhalabi et al. (2021)* proposed a smartphone-based handwritten circuit simulation system, and they relied on manual input of component values. Additionally, their system failed to deal with complex circuits and struggled to detect handwritten components accurately. On the other hand, despite detecting hand-written components successfully, *Rachala & Panicker (2022)* just reconstructed the hand-written schematic without any value parsing, netlist creation, solving the RLC circuit, or deploying the system in an end-to-end system. The proposed study mitigated all the limitations discussed so far, and a summary is presented in Table 17. As mentioned there, this study can accurately detect components, extract values automatically, process the topology, and finally produce a netlist with solutions. The system was tested by an evaluation set containing 50 circuit images with various difficulties and achieved 94% accuracy (95% CI [83.5–98.1%]).

# LIMITATIONS

Since the primary goal of this work is to establish a fully automated end-to-end pipeline from hand-drawn circuits to simulation, the current study focuses on a controlled problem setting. Several limitations remain:

- The proposed framework currently supports first-order RLC circuits with independent sources. More complex circuits with operational amplifiers or transistors are not yet supported.
- The dataset does not cover the full range of real-world drawing styles or diagram complexities.
- In noisy or overlapping sketches, errors in wire detection or node clustering can propagate to incorrect topology recovery.
- The current implementation processes each circuit in about 2 s and may require optimization for real-time deployment.

In the future, we aim to expand this work on the dataset, topology extraction robustness, and extend the framework to more complex circuit types.

# DISCUSSION

This section discusses how the data, methods, and findings of this research advance knowledge, open new questions, and help other researchers perform better work. Furthermore, how the proposed study can directly influence real-life users is also mentioned below.

## Research implications

The curated dataset of 1,000 hand-written circuit diagrams and their later versions, annotated for six class components, text regions, and 3,000 text samples with 68 engineering symbol charts, serves as a valuable resource for training and evaluation. This addresses the scarcity of handwritten engineering datasets noted in prior studies. Researchers can utilize these datasets for fine-tuning models for related tasks such as symbol recognition in other technical drawings. It also facilitates transfer learning. Pretrained models on this dataset would require less data for domain-specific adaptation.

The proposed pipeline offers a schema for hybrid systems that researchers can dissect, offering numerous research opportunities. For example, they can optimize dual detectors (YOLOv8) *via* transformer concepts or supplant classical techniques (DBSCAN) *via* a GNN (*Kipf & Welling, 2016*) to process circuit topology and determine how the system reacts. Moreover, the detailed ablation study on YOLOv8 variants across different hyperparameters provides reproducible guidance that enhances fine-tuning in resource-constrained settings and enables a meta-learning framework (*Finn, Abbeel & Levine, 2017*) to automate such a process. Finally, the human-in-the-loop web interface allows researchers to investigate user-AI workflows, which could extend to an error correction interface, quantify how manual edits reduce error in simulations (*Kohavi et al., 2009*) or integrate with federated learning for crowdsourced dataset refinement (*McMahan et al., 2017*).

### Practical implications

For educational stakeholders, the proposed system serves as an interactive tool. Using the proposed framework, students from underprivileged areas can easily deepen their understanding of electric circuit principles by comparing their hand-calculated results with the original results, which will help them identify mistakes more easily. Observing how voltage and current ratings vary with a component's value will allow them to build impulses quickly, and theorems such as superposition, Venin\Norton, or other rules will become clearer when they can easily simulate a circuit, which in turn helps them prepare for homework or exams. Teachers can also obtain real-time feedback on assignments. Not only education, but also professionals can benefit from it. The proposed system eliminates the need for manual recreation of diagrams in software, thereby reducing both time and errors. As a result, professionals or engineers can perform rapid iterations over prototyping, reviewing, or troubleshooting to successfully deduce efficient systems.

## CONCLUSION AND FUTURE WORK

This article introduces a computer-vision-based complete sketch-to-solution framework to automatically solve RLC DC circuits. To the best of our knowledge, this is one of the first fully automated approaches for 1$^{st}$ order RLC DC circuit solution. The approach combines object detection models (YOLOv8-small), OCR (CRNN + CTC), and several classical techniques (wiring/thresholding, skeletonization, DBSCAN node clustering) to identify components, extract component values, and recover the topology from the circuit diagram. In the end, it produces a SPICE-compatible netlist for circuit simulation and also performs nodal analysis to produce the current and voltage ratings across the nodes. The ablation study across YOLOv8 nano/small/medium offers practitioner-oriented guidance for future model development and the lightweight web interface for end users. The study also offers a novel dataset of 1,000 hand-written circuit diagrams. Moreover, the datasets' annotated versions for the electric components, text region of labels, and OCR dataset of engineering symbols are also curated, which will accelerate subsequent research in a similar domain.

The system achieves 98.9% accuracy in component detection and yields 6.47% character error rate for text recognition. Finally, when it was evaluated, its success rate 94% accuracy (95% CI [83.5–98.1%]). Despite having high accuracy in automation, there are several scopes for improvement. The current system mainly focuses on passive 1st-order RLC elements and independent sources. An expanded coverage of diverse elements can be explored to encompass greater diversity. The circuit topology extraction can further be refined. Beyond only providing a steady state DC solution, developing a system to provide AC and transient analysis would broaden the system's applicability.

The annotated circuit dataset and the implementation of the proposed pipeline are publicly available *via* Zenodo: https://doi.org/10.5281/zenodo.19742065. This enables full reproducibility and supports further research in this area.

## ACKNOWLEDGEMENTS

During the preparation of this work, the authors used Grammarly Premium and ChatGPT in order to improve the sentence structure, coherence, readability, and language of the manuscript. After using this tool/service, the authors reviewed and edited the content as needed and take full responsibility for the content of the published article.

## ADDITIONAL INFORMATION AND DECLARATIONS

### Funding

The authors received no funding for this work.

### Competing Interests

The authors declare that they have no competing interests.

### Author Contributions

- Mohammad Kawsar conceived and designed the experiments, performed the experiments, analyzed the data, performed the computation work, prepared figures and/or tables, authored or reviewed drafts of the article, and approved the final draft.
- Al Amin Islam Ridoy conceived and designed the experiments, performed the experiments, analyzed the data, performed the computation work, prepared figures and/or tables, authored or reviewed drafts of the article, and approved the final draft.
- Md. Rakib Hossain performed the experiments, analyzed the data, performed the computation work, authored or reviewed drafts of the article, and approved the final draft.
- Farzana Akter performed the experiments, analyzed the data, performed the computation work, authored or reviewed drafts of the article, and approved the final draft.
- AKM Bahalul Haque conceived and designed the experiments, performed the experiments, analyzed the data, performed the computation work, authored or reviewed drafts of the article, and approved the final draft.
- Rejwan Bin Sulaiman conceived and designed the experiments, performed the experiments, analyzed the data, performed the computation work, authored or reviewed drafts of the article, and approved the final draft.
- Mohammed Amin Almaiah conceived and designed the experiments, performed the experiments, analyzed the data, performed the computation work, authored or reviewed drafts of the article, and approved the final draft.

### Data Availability

The following information was supplied regarding data availability:

The code is available in the Supplemental Files.

The code is available at Zenodo: Mohammad Kawsar, & Al Amin Islam Ridoy. (2026). munsbk/Sketch2Sim-dataset-code: Initial release of Sketch2Sim (Version v1) [Computer software]. Zenodo. https://doi.org/10.5281/zenodo.19742065.

The Circuit1K data is available at Kaggle: https://www.kaggle.com/datasets/mohammadkawsar/circuit1k.

The data is available at Kaggle: https://www.kaggle.com/datasets/mohammadkawsar/yolo-component-data.

The CRNN data is available at Kaggle: https://www.kaggle.com/datasets/mohammadkawsar/crnn-data2.

The YOLO Text Detection data is available at Kaggle: https://www.kaggle.com/datasets/mohammadkawsar/yolo-text-detection-data2.

## Supplemental Information

Supplemental information for this article can be found online at http://dx.doi.org/10.7717/peerj-cs.4103#supplemental-information.